\documentclass[runningheads]{llncs}

\usepackage{eccv}

\usepackage{eccvabbrv}

\usepackage{lipsum}

\definecolor{tabhighlight}{HTML}{e5e5e5}
\definecolor{transparent}{cmyk}{0,0,0,0}
\usepackage{xcolor,colortbl}

\newcommand{\hgreen}[1]{\textcolor{ForestGreen}{\textbf{#1}}} % highlight color
 \newcommand{\hred}[1]{\textcolor{Red}{\textbf{#1}}} % highlight color

\definecolor{purple}{rgb}{0.65,0,0.65}
\definecolor{dark_green}{rgb}{0, 0.5, 0}
\definecolor{blueish}{rgb}{0.0, 0.3, .6}
\definecolor{LightCyan}{rgb}{0.88,0.95,1}

\usepackage{multirow} 

\definecolor{demphcolor}{RGB}{144, 144, 144}
\newcommand{\demph}[1]{\textcolor{demphcolor}{#1}}
\definecolor{tabhighlight}{rgb}{0.88,0.95,1}

\usepackage[font=small,labelfont=bf]{caption}

\newcommand{\method}{KDPL\xspace}
\newcommand{\CAmethod}{CA-KDPL\xspace}

\newcommand{\extmethod}{Knowledge Distillation Prompt Learning\xspace}
\newcommand{\RNum}[1]{\lowercase\expandafter{\romannumeral #1\relax}}

\newcommand{\minisection}[1]{\vspace{0.05in} \noindent \textbf{#1}.}

\usepackage{graphicx}
\usepackage{booktabs}

\usepackage[accsupp]{axessibility}  % Improves PDF readability for those with disabilities.

\usepackage{hyperref}

\usepackage{orcidlink}

\newcounter{asection}
\newcounter{asubsection}[asection]

\renewcommand{\theasection}{\Alph{asection}.}
\renewcommand{\theasubsection}{\theasection\arabic{asubsection}.}

\newcommand{\asection}[1]{
  \refstepcounter{asection}
  \section*{\theasection\ #1}
  \addcontentsline{toc}{section}{\theasection\ #1}}
\newcommand{\asubsection}[1]{
  \refstepcounter{asubsection}
  \subsection*{\theasubsection\ #1}
  \addcontentsline{toc}{subsection}{\theasubsection\ #1}}

\begin{document}

\title{Improving Zero-shot Generalization of Learned Prompts via Unsupervised Knowledge Distillation}

\titlerunning{Improving Learned Prompts via Unsupervised Knowledge Distillation}

\def\thefootnote{*}\footnotetext{These authors contributed equally to this work.}
\def\thefootnote{\arabic{footnote}}

\author{$^*$Marco Mistretta\inst{1}\orcidlink{0009-0006-6630-6477} \and
$^*$Alberto Baldrati\inst{1,2}\orcidlink{0000-0002-5012-5800} \and \\ 
Marco Bertini\inst{1}\orcidlink{0000-0002-1364-218X} \and
Andrew D. Bagdanov\inst{1}\orcidlink{0000-0001-6408-7043}
}

\authorrunning{M.~Mistretta \emph{et al.}}

\institute{University of Florence - Media Integration and Communication Center (MICC) \and
University of Pisa \\
Florence, Italy - Pisa, Italy \\
\email{\{name.surname\}@unifi.it}}

\maketitle

\begin{abstract}
Vision-Language Models (VLMs) demonstrate remarkable zero-shot generalization to unseen tasks, but fall short of the performance of supervised methods in generalizing to downstream tasks with limited data. Prompt learning is emerging as a parameter-efficient method for adapting VLMs, but state-of-the-art approaches require annotated samples. In this paper we propose a novel approach to prompt learning based on unsupervised knowledge distillation from more powerful models.  
Our approach, which we call \extmethod (\method), can be integrated into existing prompt learning techniques and eliminates the need for labeled examples during adaptation. Our experiments on more than ten standard benchmark datasets demonstrate that \method is very effective at improving generalization of learned prompts for zero-shot domain generalization, zero-shot cross-dataset generalization, and zero-shot base-to-novel class generalization problems. \method requires no ground-truth labels for adaptation, and moreover we show that even in the absence of \textit{any} knowledge of training class names \mbox{(\CAmethod)} can be used to effectively transfer knowledge. The code is publicly available at \small{\href{https://github.com/miccunifi/KDPL}{\url{https://github.com/miccunifi/KDPL}}}.
. 

\end{abstract}

\section{Introduction}\label{sec:intro}

\begin{figure*}[t]
    \centering
    \includegraphics[width=1\linewidth]{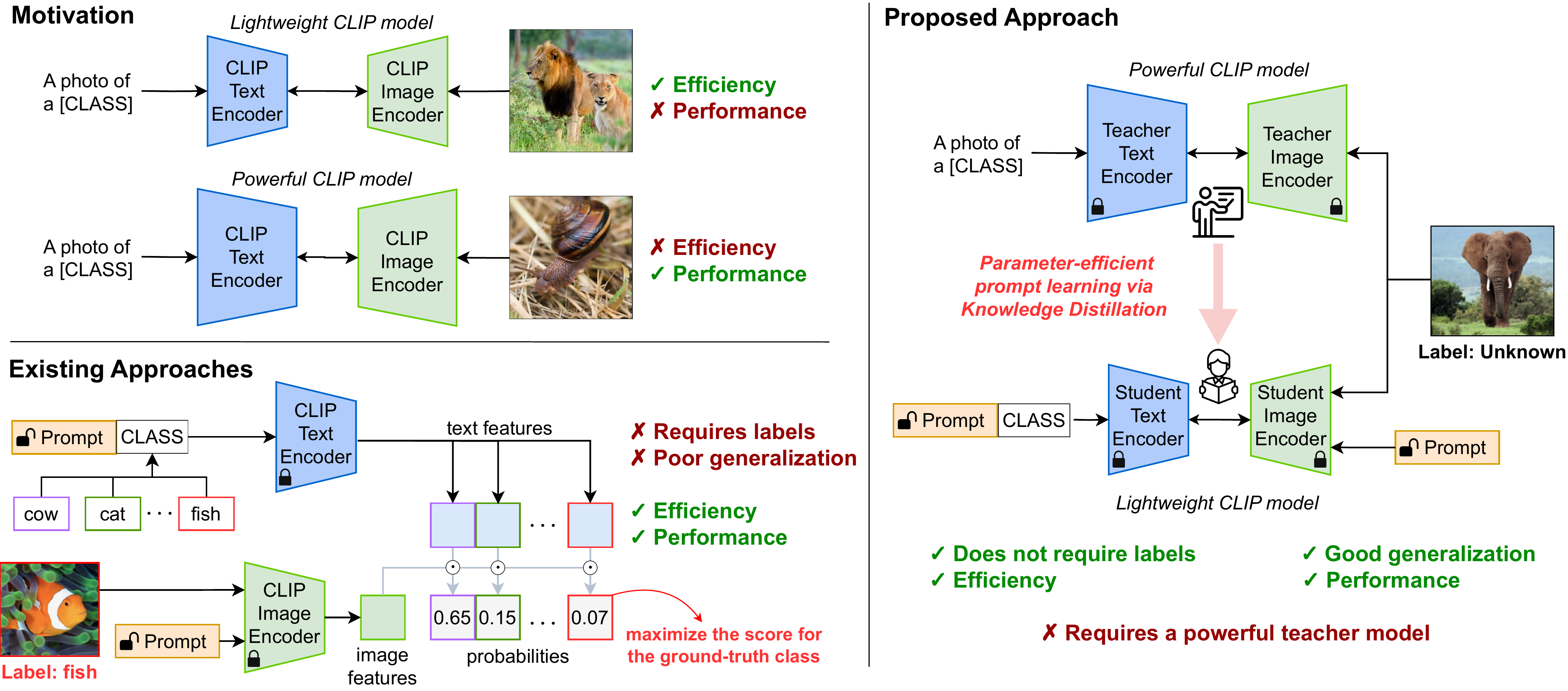}
    \caption{\textbf{Motivation and overview}. \textbf{(Top left)} Lightweight VLMs like CLIP achieve impressive zero-shot performance but lag behind supervised approaches; large VLMs incur a high computational burden. \textbf{(Bottom left)} Parameter-efficient prompt learning offers a non-destructive approach to adapting VLMs to downstream tasks; however, existing methods require annotated samples and struggle to generalize to unseen classes. \textbf{(Right)} Our approach does not require labeled samples and learns by distilling knowledge from a more powerful VLM. It can be seamlessly integrated into existing prompt learning techniques and generalizes better to unseen classes on downstream tasks. }
\label{fig:teaser}
\end{figure*}

Vision-Language Models (VLMs) like Contrastive Language-Image Pre-training (CLIP) are remarkably effective at zero-shot generalization to downstream tasks \cite{radford2021learning, zhou2022extract, baldrati2023zero, parelli2023clip}. These models leverage a dual encoder architecture and are trained to align image and text features in a shared embedding space. CLIP and similar models are able to perform zero-shot classification by predicting the output class based on the similarity between the test image embedding and text embeddings of words from a fixed vocabulary.

While CLIP exhibits remarkable zero-shot performance, it can fall short of supervised adaptation. Narrowing this gap through fine-tuning is challenging when large training datasets are not available for downstream tasks. Moreover, fine-tuning can also lead to overfitting and is destructive in that it can cause the model to forget knowledge acquired during large-scale pre-training~\cite{ding2022don, zhou2022learning}.

An intuitive way to achieve better performance is to use larger and more powerful VLMs~\cite{cherti2023reproducible} with better zero-shot generalization. However, though performance improves with increasing size, their practical applicability in real-world scenarios decreases. Lightweight CLIP models based on ResNet-50 and ViT-B/32 require about 18 and 14 GFLOPs for inference, respectively. In contrast, more powerful models based on ViT-H/14~\cite{fang2023data, cherti2023reproducible} require nearly $30\times$ the computation at 381 GFLOPs~\cite{fang2023data} for inference.

To address these challenges, parameter-efficient prompt learning is emerging as a promising and non-destructive approach to adapting VLMs to downstream tasks~\cite{zhong2021factual, lester2021power, li2021prefix}. Zhou et al. \cite{zhou2022learning} proposed a text-only prompt learning approach called CoOp to overcome the limitations of manually crafted prompts. Subsequently, Jia et al. \cite{jia2022visual} extended prompt learning to the visual domain with Visual Prompt Tuning (VPT), and Khattak et al.~\cite{khattak2023maple, khattak2023self} proposed a multi-modal prompt learning approach called MaPLe~\cite{khattak2023maple} and a self-regulated approach called PromptSRC~\cite{khattak2023self}. Although significantly improving over carefully tuned, hand-crafted prompts, these state-of-the-art techniques all require annotated samples from the dataset and have problems generalizing to other datasets or classes unseen during training~\cite{zhou2022conditional, khattak2023maple}.

To eliminate the need for labeled training examples and improve the generalization of learned prompts, we propose a novel approach to prompt learning which we call \extmethod (\method). \method adapts lightweight VLMs and improves performance on downstream tasks by distilling knowledge from a more powerful VLM without the need for annotated examples. \method is a flexible method that can be seamlessly integrated into existing prompt learning framework such as CoOp~\cite{zhou2022learning}, CoCoOp~\cite{zhou2022conditional}, VPT~\cite{jia2022visual}, MaPLe~\cite{khattak2023maple}, and PromptSRC~\cite{khattak2023self}. \Cref{fig:teaser} summarizes the motivations and illustrates the workflow of the proposed approach. We validate \method on two increasingly challenging scenarios: in the first, we assume knowledge of training class names, while in the second we assume no knowledge of training classes. Importantly, in both scenarios \emph{no annotated training samples are used.}

The contributions of this work are:
\begin{itemize}
    \item we propose a novel parameter-efficient prompt-learning approach that eliminates the need for labeled training examples by distilling knowledge from a large Vision-Language Model;
    \item we show that our approach can be integrated into existing prompt learning techniques to learn visual, textual, and multimodal prompts which generalize significantly better than state-of-the-art baselines to unseen classes in downstream tasks;
    \item we demonstrate the effectiveness of our approach through extensive experiments on more than ten standard benchmark datasets covering a broad range of downstream problems, including domain generalization, cross-dataset adaptation, and base-to-novel generalization; and

    \item we introduce a new, class agnostic evaluation setting in which training class names are unknown and show the superiority of our approach in this scenario.

\end{itemize}
To the best of our knowledge ours is the first approach to parameter-efficient VLM adaptation which is applicable in scenarios that are \emph{label agnostic} (i.e.~no label supervision is required during adaptation) and also in scenarios that are additionally \emph{class agnostic} (i.e.~no knowledge of training class names is assumed).  

\section{Related Work}
In this section we review the recent literature most relevant to our contribution.

\minisection{Vision-Language Models}
Vision-Language Models (VLMs) learn robust visual representations thanks to their extensive pre-training on image-caption datasets~\cite{schuhmann2021laion,schuhmann2022laion}. These representations are very effective at generalizing to a variety of downstream tasks~\cite{desai2021virtex,radford2021learning, jia2021scaling}. In contrast to vision models trained only with image supervision, vision-language models can interpret both visual and textual data, showing improvement in tasks requiring cross-model understanding~\cite{baldrati2023zero, barraco2022unreasonable, song2022clip}.
We focus on contrastively-trained VLMs such as CLIP~\cite{radford2021learning}, ALIGN~\cite{jia2021scaling}, and LiT~\cite{zhai2022lit}. These models use a dual encoder architecture and contrastive learning to align images and text in a common embedding space. Thanks to this shared embedding space and massive pre-training datasets
these models achieve remarkable performance. However, there is still a gap between the zero-shot capabilities of these VLMs and the performance of tailored state-of-the-art models~\cite{radford2021learning, dong2022clip}, and thus there is active research on efficient and effective methods to adapt VLMs to downstream tasks~\cite{zhang2021tip, wang2023improving, gu2021open, luddecke2022image}.

\minisection{Prompt Learning}
The performance of VLMs is highly dependent on textual prompts used to characterize downstream tasks. Given the significant impact of small changes in wording on performance, the manual generation of optimal textual prompts is a challenging task. Inspired by developments in NLP~\cite{zhong2021factual, lester2021power, li2021prefix}, CoOp~\cite{zhou2022learning} proposes to learn continuous vectors in the word embedding space instead of using tailored hand-crafted prompts.

The promising results of CoOp have attracted considerable interest in prompt learning for VLMs~\cite{lu2022prompt, khattak2023maple, yao2023visual, agnolucci2023eco, bulat2023lasp, chen2022plot, khattak2023self, shi2023logoprompt, zhu2023prompt, shi2024dept, abdul2024align}. CoCoOp, in particular, highlights the poor performance of CoOp on unseen classes and uses image-conditional dynamic prompts to improve generalization~\cite{zhou2022conditional}. PLOT~\cite{chen2022plot} instead is based on learning multiple prompts by exploiting local visual features. KgCoOp~\cite{yao2023visual} adds a regularization loss to minimize the discrepancy between learned and hand-crafted prompts, while VPT~\cite{jia2022visual} adapts prompt learning to the visual domain by learning continuous vectors in the input space of a ViT~\cite{dosovitskiy2020image}. Finally, MaPLe~\cite{khattak2023maple} introduces a multimodal prompt learning strategy for VLMs. 
PromptSRC~\cite{khattak2023self} introduces a self-regularization framework for prompting, addressing overfitting through agreement with a frozen model, a self-ensemble of prompts, and textual diversity.

Similar to the approach we propose in this paper, UPL~\cite{huang2022unsupervised} performs prompt learning without relying on annotated samples. It leverages pseudolabels derived from a larger CLIP model and selects the top-K most confident samples per class to construct a balanced set of pseudo-labels. UPL requires a substantial collection of unlabeled images and is not directly applicable when only a few unlabeled samples are available, like in the few-shot adaptation scenarios considered here. In contrast to UPL, \method does not use pseudolabeling; instead, we directly learn from the logits of a CLIP teacher model via knowledge distillation, thus eliminating the need for any selection strategy of the training samples.

\minisection{Knowledge Distillation}
Knowledge distillation (KD) is a machine learning technique in which a simple model (\textit{student}) is trained to mimic the behavior of a larger model (\textit{teacher}) by learning from its output or intermediate activation~\cite{hinton2015distilling}. This technique has found success in many contexts, including image classification~\cite{hinton2015distilling, beyer2022knowledge, chen2019knowledge}, self-supervised learning~\cite{xu2020knowledge, fang2020seed}, and image/video segmentation~\cite{he2019knowledge, mullapudi2019online, dou2020unpaired, siam2019video}, leading to improvements in model compression, computational efficiency, and performance.
DML~\cite{zhang2018deep} introduces a mutual learning approach to train students and teachers simultaneously. DKD~\cite{zhao2022decoupled} reformulates the classical KD loss into a term related to the target class and a term related to the non-target classes. In contrast to the majority of knowledge distillation approaches~\cite{zhang2018deep, zhao2022decoupled, yun2020regularizing, yuan2020revisiting}, we do not utilize any labels during training. Instead, we solely employ the distillation loss. In our work, we apply knowledge distillation in a parameter-efficient prompt learning scenario. Specifically, we distill knowledge from a large and powerful VLM (teacher) into a lightweight VLM (student) by updating only a reduced number of parameters (the prompt).

\minisection{Techniques Using Teacher-Student Distillation}
Recently Li et al.~\cite{li2024promptkd} introduced PromptKD which also uses teacher-student distillation in prompt learning. Although it too leverages a larger teacher to guide a smaller student, it differs from our contribution in several key aspects. PromptKD uses PromptSRC~\cite{khattak2023self} to pre-train the teacher model with labeled examples, whereas \method does not pre-train the teacher and requires no labeled examples. PromptKD uses an MLP to project the student image encoder output, which adds extra parameters. \method is instead a general framework with no additional parameters, adaptable to any prompt-tuning technique to render them completely unsupervised. Moreover, PromptKD uses the entire training set, including both base and novel images, for distillation. In contrast, \method follows the standard 16-shot setting. 

\section{Knowledge Distillation Prompt Learning (\method)}
\label{sec:method}
We first introduce preliminary concepts related to prompt learning and then describe our approach to applying knowledge distillation to the problem.

\subsection{Preliminaries}
Our approach is based on knowledge distillation applied to prompt learning. Here we discuss the base teacher and student models (CLIPs) and five state-of-the-art prompt learning approaches into which we will incorporate \method.

\minisection{CLIP}
Contrastive Language-Image Pre-training (CLIP) is a vision-language model trained to align images and textual captions in shared semantic space~\cite{radford2021learning}. CLIP consists of an image encoder $f_{\theta}$ and a text encoder $g_{\phi}$. Given an image $I$, the image encoder computes its feature representation $f_{\theta}(I) \in \mathbb{R}^{d}$, where $d$ is the size of the semantic embedding space. Similarly, for a given textual caption $Y$, a word embedding layer $E_L$ maps each tokenized word to the token embedding space $\mathcal{W}$. Then, the text encoder $g_{\phi}$ generates the textual feature representation $g_{\phi}(E_L(Y)) \in \mathbb{R}^{d}$. The main goal of CLIP training is to learn $\theta$ and $\phi$ such that $f_{\theta}(I)\approx g_{\phi}(E_L(Y))$ for associated image/text pairs $(I,Y)$.

When using a Vision Transformer (ViT)~\cite{dosovitskiy2020image} as the visual encoder $f_{\theta}$, the encoding process begins by splitting the image into $U$ fixed-size patches. These patches are then projected into patch embeddings $\{ w_1, w_2, \ldots, w_U \}$, where each $w_i$ belongs to the patch embedding space $\mathcal{V}$. A learnable class (CLS) token $c_i$ is concatenated with the patch embeddings, resulting in the input to the vision transformer being $\{c_i, w_1, w_2, \ldots, w_U \}$. Finally, the CLS token of the final transformer layer is projected to the shared embedding space via a linear projection to obtain the final representation.

To perform zero-shot classification using CLIP, we start with an image $I$ and build a set of textual prompts $\{Y_i\}_{i=1}^C$, where $C$ denotes the number of classes. Each handcrafted text prompt $Y_i$ takes the format  ``\textit{a photo of a [$\text{CLASS}_i$]}'', where $\text{CLASS}_i$ represents a specific class name, such as \textit{airplane}, \textit{bird}, etc. 

Then the feature representations $\psi_I = f_{\theta}(I)$ and $\psi_T^i = g_{\phi}(E_L(Y_i))$ are extracted using the CLIP encoders. The predicted probability for each class is:
\begin{equation} \label{eq:prob}
p(y=i|I) = \frac{\exp( \cos(\psi_T^i, \psi_I)/\tau) }{\sum_{j=1}^C \exp( \cos(\psi_T^j, \psi_I)/\tau)},
\end{equation}
where $\cos(\cdot, \cdot)$ is the cosine similarity and $\tau$ is a temperature hyperparameter.

\minisection{Prompt-Learning Techniques}
Here we summarize how the state-of-the-art techniques for textual, visual, and multimodal prompt learning work. 
\begin{itemize}
\item \textbf{CoOp}~\cite{zhou2022learning} is a textual prompt learning technique that learns continuous context tokens (i.e. the learnable prompt) in the CLIP token embedding space. Specifically, CoOp introduces $M$ learnable vectors, $\{ v_1, v_2, \ldots, v_M \}$ where each context vector $v_i \in \mathcal{W}$. For each of the $k$ classes of a dataset, the input to the text encoder is $\{ v_1, v_2, \ldots, v_M, c_k \}$, where $c_k = E_L([CLASS_k])$.

\item \textbf{CoCoOp}~\cite{zhou2022conditional} extends CoOp by incorporating a lightweight network $h_{\theta}$. Each context token is obtained as $v_i(I) = v_i + \pi$, where $\pi = h_{\theta}(I)$. This method ensures that the textual prompts are conditioned on the input image.

\item \textbf{VPT}~\cite{jia2022visual} is a visual prompt learning method that can be viewed as the counterpart of CoOp in the visual domain. Unlike CoOp, which operates entirely in the textual token embedding space $\mathcal{W}$, VPT performs prompt learning in the visual patch embedding space $\mathcal{V}$. Specifically, VPT learns $P$ visual tokens, leading to the input to the ViT being $\{ z_1,\ldots, z_P,c_i, w_1, w_2, \ldots, w_U \}$. VPT offers two prompting variants: deep and shallow. The deep variant learns a distinct prompt for each ViT layer, while the shallow variant incorporates the prompt parameters only into the input of the first layer.

\item \textbf{MaPLe}~\cite{khattak2023maple} is a deep multi-modal prompt learning technique that promotes strong coupling between vision and language prompts. In practice, MaPLe learns different textual context tokens for each layer in $g_{\phi}$. The visual prompts, on the other hand, are not directly learned but are obtained through a linear mapping from the textual ones.

\item \textbf{PromptSRC}~\cite{khattak2023self} is another multi-modal prompt learning technique that optimizes prompts through mutual agreement maximization between prompted and frozen model features using self-distillation, Gaussian-weighted prompt aggregation over the training session, and promoting textual diversity to address sample diversity imbalance.

\end{itemize}
Note that methods involving visual prompt learnable tokens like VPT~\cite{jia2022visual}, MaPLe~\cite{khattak2023maple}, and PromptSRC~\cite{khattak2023self} can only be applied to VLMs equipped with a ViT-based image encoder.

\subsection{Label Agnostic Prompt Learning}\label{sec:label-agnostic}
The methods described above all rely on ground-truth labels during adaptation. Here we show how unsupervised knowledge distillation can be used instead to replace the need for annotated training examples.

\minisection{Overview}
Our proposed approach, which we call Knowledge Distillation Prompt Learning (\method), is a general method designed to enhance the performance of the CLIP model on downstream tasks through parameter-efficient prompt learning. Unlike previous approaches~\cite{zhou2022conditional, zhou2022learning, khattak2023maple}, which rely on labeled examples for training, \method eliminates the need for manually-labeled samples by learning only through knowledge distillation from a larger and more powerful VLM. Note that \method is a method that can be seamlessly integrated with any existing prompt learning approach in scenarios where no information about class names or labels is available.

We validate \method in two progressively challenging supervision regimes. In the \textbf{label agnostic} scenario we do not use ground-truth labels, but we assume knowledge of the class names in the training dataset. In the \textbf{class agnostic} scenario (see Section~\ref{sec:class-agnostic}) we go one step further and assume that even the training class names are unknown. For this class agnostic scenario we propose an effective and efficient online strategy for automatically filtering the classes from a large dictionary of approximately 20K class names~\cite{kuznetsova2020open}.

\begin{figure*}[t]
    \centering
    \includegraphics[width=0.98\linewidth]{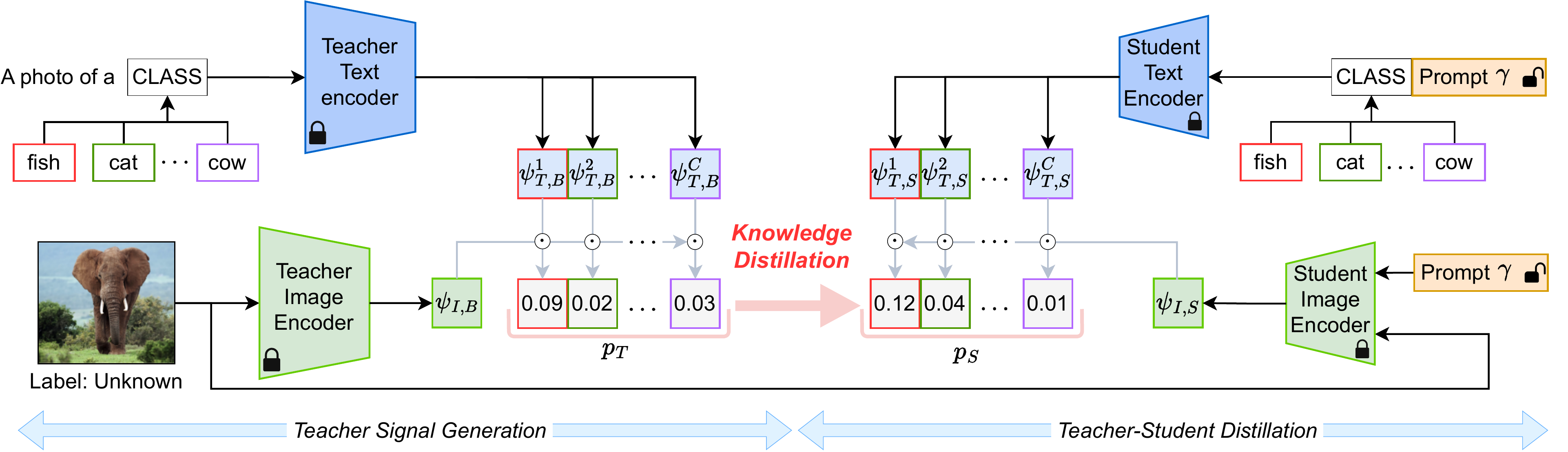}
    \caption{
     \textbf{\extmethod (\method) overview}. Given a lightweight VLM \textit{student} and a larger, more powerful VLM \textit{teacher}, \method updates the student prompt parameters by distilling knowledge from the teacher. \method first performs zero-shot classification with the teacher to obtain teacher probabilities $p_T$. It then computes the student probabilities $p_S$ and performs knowledge distillation to update the student prompt parameters $\gamma$.}
   \label{fig:method}
\end{figure*}

\minisection{Prompt Learning via Unsupervised Knowledge Distillation (KDPL)}
Given a lightweight CLIP model (the \textit{student}) and a larger, more powerful CLIP model (the \textit{teacher}), we aim to improve the downstream performance of the student model by distilling knowledge from teacher to student. For an image $I$ and a set of classes $\mathcal{C} = \{ \text{CLASS}_i \}_{i=1}^{C}$, we start by performing zero-shot classification using the frozen teacher model. Specifically, we use the teacher image encoder $f_{\theta}^B$ and text encoder $g_{\phi}^B$ to compute the teacher image features $\psi_{I,B} = f_{\theta}^B(I)$ and text features $\psi_{T, B}^i = g_{\phi}^S(E_L([\text{CLASS}_i]))$. For the teacher model we use the fixed text prompt ``a photo of [CLASS]''. We then apply \cref{eq:prob} to produce the probabilities $p_T(I, \mathcal{C})$ predicted by the teacher on image $I$ for classes $\mathcal{C}$.

The teacher model does not rely on a learnable prompt and its predictions remain fixed during training. Our aim is to learn text and image prompts for the student model that enhance its generalization to downstream tasks.

\newpage
We denote with $f_{\theta}^S$ and $g_{\phi}^S$ the student image and text encoders, respectively, and with $\gamma$ the parameters associated with the learnable student prompts (see Fig.~\ref{fig:method}). Given the same image $I$ processed by the teacher and the same set of classes $\mathcal{C}$, the student extracts image features $\psi_{I,S} = f_{\theta, \gamma}^S(I)$ and text features $\psi_{T, S}^i = g_{\phi,\gamma}^S(E_L([\text{CLASS}_i]))$. Note that the text and image encoders can both depend on the prompt parameters $\gamma$. According to the prompt learning technique used, the $\gamma$ parameters can be used only by the text encoder (CoOp, CoCoOp), the visual encoder (VPT), or both (MaPLe, PromptSRC). Finally, using \cref{eq:prob} we produce student class probabilities $p_S(I, \mathcal{C})$ predicted on image $I$ for classes $\mathcal{C}$. Note that all encoder parameters except for the learnable prompt $\gamma$ are frozen.

We use the symmetric KL-divergence between the teacher ($p_T(I, \mathcal{C})$) and the student ($p_S(I, \mathcal{C})$) probabilities in a distillation loss:
\begin{equation}
\label{eq:loss}
    \mathcal{L(I, \mathcal{C})} = \left[ D_{\text{KL}}(p_T(I, \mathcal{C}) \, || \, p_S(I, \mathcal{C}))  + D_{\text{KL}}(p_S(I,\mathcal{C}) \, || \, p_T(I,\mathcal{C})) \right],
\end{equation}
where $D(q \, || \, p)$ is the asymmetric KL-divergence between the discrete probability distributions $p$ and $q$:
\begin{align}
D_{\text{KL}}(p \, || \, q) = \sum_{i \in \mathcal{C}} p_i \log \left( \frac{p_i}{q_i} \right)
\end{align}
This distillation loss depends only on the fixed predictions of the teacher, the prompt-conditioned predictions of the students, and the set of classes $\mathcal{C}$. 

Importantly, \textit{our distillation loss does not assume any knowledge of class labels for image $I$}. Nor does it require that $\mathcal{C}$ be the classes of the downstream task -- that is, \method can be used for Label Agnostic and Class Agnostic adaptation scenarios. We found in early experiments that the symmetric KL-divergence works slightly better than either asymmetric option (see Section \ref{supp:symmetric} in the Supplementary Material for an ablation study on this choice).

\subsection{Class Agnostic Prompt Learning}\label{sec:class-agnostic}
To further evaluate the generalization capabilities of \method, we introduce a scenario where not only do we not know the \textit{labels} of training images (label agnostic, Section~\ref{sec:label-agnostic}) but where we also do not even know the \textit{class names} associated with the dataset (class agnostic). This scenario is considerably more challenging as we make no assumptions about the few-shot training data.

To address the unavailability of class names, we propose a strategy for automatically selecting a set of class names for each batch. We start with a large vocabulary of class names from which to select. Specifically, we use the Open Images V7 dataset~\cite{kuznetsova2020open}, which contains $\sim$20K classes. The most straightforward method would simply use all 20K classes during training. However, this is impractical as the memory required by prompt learning methods increases linearly with the number of classes. According to Ren et al.~\cite{ren2024prompt}, CoOp requires nearly 15MB of VRAM \textit{per class}, resulting in approximately 300GB of memory when multiplied by 20K classes. Therefore, we propose a method to automatically select which classes to use for each batch by retaining only the ones most relevant to the calculation of the loss in~\cref{eq:loss}.

Given a batch of images $X=\{ I_i \}_{i=1}^N$ and all the class names in the vocabulary $\mathcal{C} = \{ \text{CLASS}_i \}_{i=1}^{C}$, where $N$ represents the number of images in a batch and $C$ the size of the dictionary, we let the teacher model select the most relevant classes. Our objective is to identify the most useful $K$ classes in each batch for student prompt learning. After extracting the teacher image and text features, for each image $I_i \in X$, we apply \cref{eq:prob} to obtain the teacher probabilities $p_{T}(I_i, \mathcal{C})$. By stacking the probabilities along the batch dimension, we obtain the matrix $P_T = \left[p_{T}(I_1, \mathcal{C});\, \cdots; \, p_{T}(I_N, \mathcal{C} ) \right]^T \in \mathbb{R}^{N \times C}$, where the $i$-th row corresponds to the probabilities associated with image $I_i$. We then compute the average probabilities along the batch axis, resulting in $\overline{P_T} \in \mathbb{R}^{C}$. Finally, we select the classes corresponding to the $K$ highest values in $\overline{P_T}$.

Using the teacher model to perform this class filtering for each batch is feasible and does not incur excessive memory costs since the teacher requires no gradient computation. Therefore, the memory consumption does not depend on the number of classes in $\mathcal{C}$. Conversely, although the student model remains frozen, gradients must still be propagated to update the prompt parameters $\gamma$. Once the classes are selected, the training strategy remains the same as described above, with the only difference being that the class names observed by the student vary in each batch based on teacher predictions.

\section{Experimental Results}
\label{sec:experiments}
In this section we report on experiments validating our proposed approach. 

\subsection{Evaluated Scenarios and Implementation Details}
Following previous works~\cite{zhou2022conditional, zhou2022learning, khattak2023maple}, we validate \method in three distinct settings: 1) domain generalization; 2) cross-dataset transfer; 3) generalization to unseen classes. Additionally, to evaluate the scenario where class names are also unknown, we introduce a new evaluation setting we call \emph{class agnostic adaptation}. We use the train/val/test splits and seeds provided by Zhou et al.~\cite{zhou2022learning} for all datasets.\footnote{\href{https://github.com/KaiyangZhou/CoOp/blob/main/DATASETS.md}{\url{https://github.com/KaiyangZhou/CoOp/blob/main/DATASETS.md}}} All reported results are averages over three independent runs.

\minisection{Evaluated Scenarios}
We evaluate \method on the following scenarios:
\begin{itemize}
\item \textbf{Domain Generalization}.
To assess the ability of the learned prompts to generalize to out-of-distribution datasets, we apply the prompt learned from ImageNet~\cite{deng2009imagenet} to four different versions of ImageNet datasets exhibiting various types of domain shift. The target datasets are ImageNetV2~\cite{recht2019imagenet}, ImageNet-Sketch~\cite{wang2019learning}, ImageNet-A~\cite{hendrycks2021natural}, and ImageNet-R~\cite{hendrycks2021many}.

\item \textbf{Cross-dataset Transfer}.
To evaluate the ability of learned prompts to generalize to unseen classes, we evaluate the same prompt trained on ImageNet on a broad range of downstream recognition tasks. We validate on ten datasets with varying characteristics: Caltech101~\cite{fei2004learning} for general object classification; OxfordPets~\cite{parkhi2012cats}, StanfordCars~\cite{krause20133d}, Flowers102~\cite{nilsback2008automated}, Food101~\cite{bossard2014food}, and FGVCAircraft~\cite{maji2013fine} for fine-grained classification; the Describable Textures Dataset (DTD)~\cite{cimpoi2014describing} for texture classification; EuroSAT~\cite{helber2019eurosat} for satellite-image recognition; and UCF101~\cite{soomro2012ucf101} for action recognition.

\item \textbf{Generalization to Unseen Classes}.
To evaluate how learned prompts generalize to unseen classes from the same dataset, we divide classes into two subsets: base and novel classes. The prompt is trained exclusively on the base classes and then tested on the novel ones. We evaluate on ImageNet as well as the ten benchmark datasets used for cross-dataset evaluation.

\item \textbf{Class Agnostic Adaptation}.
In addition we introduce a novel evaluation setting in which the training class names are unknown. The prompt is trained on ImageNet and tested on all the benchmark datasets used in the domain generalization and cross-dataset evaluations.
\end{itemize}

\minisection{Implementation Details}
We use a CLIP model with a ViT-H-14 visual backbone as the teacher model.\footnote{\href{https://huggingface.co/apple/DFN5B-CLIP-ViT-H-14}{\url{https://huggingface.co/apple/DFN5B-CLIP-ViT-H-14}}} For student models, we evaluate both on a CLIP model based on ResNet-50 and a model based on ViT-B/32. By experimenting with both ResNet-based and ViT-based CLIP models we show the architecture independence of our approach. 
To assess how \method performs when integrated into different prompt learning techniques, we selected five distinct textual, visual, and multimodal approaches. 
With the ResNet-50 student model, we experiment with CoOp~\cite{zhou2022learning} and CoCoOp~\cite{zhou2022conditional}. For the ViT-B/32 student, since it supports both visual and textual prompting, we experiment with CoOp~\cite{zhou2022learning}, VPT~\cite{jia2022visual}, MaPLe~\cite{khattak2023maple} and PromptSRC~\cite{khattak2023self}. 

We denote the integration of our unsupervised knowledge distillation strategy with a specific prompt learning technique by adding the suffix +\method to the name of the corresponding technique. For example, CoOp+\method refers to the application of \method to CoOp. For class agnostic settings we instead use the suffix +\CAmethod to indicate that no training class names are used. 

Unless otherwise stated, all experiments are conducted in the few-shot setting with 16 examples per class randomly sampled from the training set. We set the temperature hyperparameter $\tau$ in \cref{eq:prob} to $0.01$. In the class agnostic experiments in which we assume no knowledge of class names in the training set, we set the number of class names selected in each iteration to $K=1000$.

For each prompt learning method we use the original implementation and hyperparameters reported in the respective papers. Since we use the compact ResNet-50 and ViT-B/32 backbones, we cannot directly compare to the originally published results. Thus, all numbers we report here were computed by us and are averages over three independent runs. See Section \ref{supp:impl-details} in the Supplementary Material for additional implementation details.

\begin{table*}[t]
    \caption{
    \textbf{Domain Generalization}. Comparison between the baselines and the proposed unsupervised \method variants (highlighted in cyan). In this setting, the prompt is trained on ImageNet (source) and then tested on four different versions of ImageNet (targets) that exhibit some kind of domain shift. Average performance improvements over the baselines are shown in \hgreen{green}, and average performance deterioration in \hred{red}.}
    \label{tab:domain-generalization}
    \centering
    \resizebox{1\textwidth}{!}{
        \begin{tabular}{c l cccccc}
        \toprule
        & & Source & \multicolumn{5}{c}{Target}  
        \\ \cmidrule(lr){3-3} \cmidrule(lr){4-8}
        
        Backbone & Method & ImageNet & -V2 & -S & -A & -R & \textit{Average}\\  
        \midrule
         
         & \demph{CLIP (student)} & \demph{58.20} & \demph{51.50} & \demph{33.30} & \demph{21.70} & \demph{56.00} & \demph{40.63} \\
        \cmidrule(lr){2-8}
        
         & CoOp & 62.40 & 55.17 & 33.70 & 23.13 & 56.20 & 42.05\\

         \rowcolor{tabhighlight}  \cellcolor{white}  & CoOp + \method &  \textbf{62.73}  & \textbf{55.37} & \textbf{35.20} & \textbf{23.27} & \textbf{57.77} & \hgreen{42.90} \\ 
        
        \cmidrule(lr){2-8}
    
          \multirow{-4}{*}{RN50} & CoCoOp & \textbf{63.07} & 55.53 & 34.77 & \textbf{23.73} & \textbf{59.47} & 43.38\\

        \rowcolor{tabhighlight} \cellcolor{white} & CoCoOp + \method & 62.70 & \textbf{55.60} & \textbf{35.30} & 23.43 & 57.90 & \hred{43.06}\\ 
         
        \midrule
        
        & \demph{CLIP (student)} & \demph{62.00} & \demph{54.70} & \demph{40.80} & \demph{29.60} & \demph{66.00} & \demph{47.78} \\
        \cmidrule(lr){2-8}
        
         & CoOp & \textbf{66.33} & \textbf{58.30} & 41.40 & 31.47 & 65.87 & 49.26 \\
        
         \rowcolor{tabhighlight} \cellcolor{white}& CoOp + \method & 65.97 & 58.10 & \textbf{42.50} & \textbf{31.63} & \textbf{67.37} & \hgreen{49.90} \\
     
        \cmidrule(lr){2-8}
    
        & VPT & 64.97 & 56.73 & 41.27 & 27.00 & 66.50 & 47.88 \\
        
         \rowcolor{tabhighlight} \cellcolor{white} & VPT + \method & \textbf{65.10} & \textbf{57.37} & \textbf{41.67} & \textbf{27.77} & \textbf{67.47} & \hgreen{48.57}\\ 
         
        \cmidrule(lr){2-8}
        
         \multirow{-4}{*}{ViT-B/32 } & MaPLe & \textbf{66.80} & \textbf{58.53} & 42.23 & \textbf{30.13} & 66.40 & 49.32\\

        \rowcolor{tabhighlight} \cellcolor{white} & MaPLe + \method & 66.50 & 58.47 & \textbf{42.77} & 29.87 & \textbf{67.70} & \hgreen{49.70} \\  

        \cmidrule(lr){2-8}

         & PromptSRC & \textbf{66.33} & \textbf{58.70} & 42.97 & \textbf{32.07} & \textbf{68.93} & 50.67 \\
        
         \rowcolor{tabhighlight} \cellcolor{white} & PromptSRC + \method & 66.27 & 58.60 & \textbf{43.07} & 31.70 & 68.83 & \hred{50.55}\\ 
    
        \midrule
        
        ViT-H/14 & \demph{CLIP (teacher)} & \demph{82.80} & \demph{76.60} & \demph{71.10} & \demph{71.10} & \demph{91.30} & \demph{77.53}\\
    
        \bottomrule
        \end{tabular}
}
\end{table*}

\subsection{Results on Domain Generalization}
\Cref{tab:domain-generalization} outlines the performance in the domain generalization setting. We report the performance of each baseline method alongside their unsupervised \method variants. Additionally, the performance of the zero-shot CLIP student and teacher models using a handcrafted prompt is included for comparison.
We observe that applying our unsupervised approach to each baseline does not result in a significant decrease in performance on the source dataset. Notably, when transferring the learned prompts to a different domain, incorporating \method in each baseline can lead to a slight improvement. This suggests that our unsupervised teacher-student distillation learns prompts that generalize better than those trained using ground-truth labels. The only exceptions where we observe an average decrease compared to the baseline are when using CoCoOp or PromptSRC. CoCoOp-\method achieves an average accuracy significantly lower than CoCoOp only on the ImageNet-R dataset. Finally, note that all our unsupervised \method variants significantly improve the performance over the zero-shot CLIP student model in both the source and the target datasets.

\subsection{Results on Cross-dataset Transfer}
\begin{table*}[t]
    \caption{\textbf{Cross-dataset Transfer}. Comparison between the baselines and the proposed unsupervised \method variants (highlighted in cyan). The prompt is learned on ImageNet and then tested on the ten different target datasets. Average performance improvements over the baselines are indicated in \hgreen{green}.
    }
    \label{tab:cross-dataset}
    \centering
    \resizebox{\textwidth}{!}{

    \begin{tabular}{c l ccccccccccc}
    \toprule
    & & \multicolumn{11}{c}{Target} \\ \cmidrule(lr){3-13}
    Backbone &
    Method & \rotatebox{80}{OxfordPets} & \rotatebox{80}{Flowers102} & \rotatebox{80}{FGVCAircraft} & \rotatebox{80}{DTD} & \rotatebox{80}{EuroSAT} & \rotatebox{80}{StanfordCars} & \rotatebox{80}{Food101} & \rotatebox{80}{SUN397} & \rotatebox{80}{Caltech101} & \rotatebox{80}{UCF101} & \rotatebox{80}{\emph{\textit{Average}}} \\
    \midrule
    
    & \demph{CLIP (student)} & \demph{83.70} & \demph{61.00} & \demph{15.60} & \demph{40.00} & \demph{24.20} & \demph{55.60} & \demph{75.20} & \demph{58.50} & \demph{86.00} & \demph{58.30} & \demph{55.81} \\
    \cmidrule(lr){2-13}

    & CoOp & 84.60 & 61.63 & 13.77 & 36.83 & 22.33 & 54.20 & 75.03 & 59.00 & 87.67 & 57.73  & 55.28 \\

     \rowcolor{tabhighlight} \cellcolor{white} & CoOp + \method & \textbf{87.33} & \textbf{62.83} & \textbf{14.97} & \textbf{39.27} & \textbf{30.17} & \textbf{56.60} & \textbf{77.27} & \textbf{60.30} & \textbf{88.57} & \textbf{58.63} & \hgreen{57.59}\\

    \cmidrule(lr){2-13}

     \multirow{-4}{*}{RN50} & CoCoOp &  86.93 & 63.47 & \textbf{16.27} & 40.53 & 27.20 & 55.47 & 77.73 & 61.40 & 88.07 & \textbf{60.43} & 57.75\\

    \rowcolor{tabhighlight} \cellcolor{white} & CoCoOp + \method  & \textbf{87.33} & \textbf{64.97} & 15.97 & \textbf{41.27} & \textbf{29.17} & \textbf{56.83} & \textbf{78.00} & \textbf{61.50} & \textbf{88.50} & 59.53 & \hgreen{58.31}\\
 
    \midrule

     & \demph{CLIP (student)} & \demph{85.00} & \demph{64.30} & \demph{18.20} & \demph{42.80} & \demph{38.10} & \demph{60.40} & \demph{79.10} & \demph{62.00} & \demph{91.10} & \demph{60.70} & \demph{60.17} \\

    \cmidrule(lr){2-13}

    & CoOp & \textbf{87.13} & 62.20 & 12.13 & \textbf{40.57} & 36.97 & 57.83 & 80.13 & 62.83 & 91.13  & \textbf{61.80} & 59.27\\

    \rowcolor{tabhighlight} \cellcolor{white} & CoOp + \method & \textbf{87.13} & \textbf{62.23} & \textbf{15.57} & 40.40 & \textbf{40.93} & \textbf{58.83} & \textbf{80.57} & \textbf{64.23} & \textbf{92.77} & 61.63 & \hgreen{60.43}\\ 

    \cmidrule(lr){2-13}

    & VPT & \textbf{87.43} & \textbf{64.77} & \textbf{19.53} & \textbf{44.20} & 28.47 & 57.60 & 78.27 & \textbf{63.73} & 91.40  & 62.50 & 59.79\\

    \rowcolor{tabhighlight} \cellcolor{white} & VPT + \method & \textbf{87.43} & 64.43 & 17.77 & 42.93 & \textbf{30.80} & \textbf{57.97} & \textbf{78.47} & 63.63 & \textbf{92.00}  & \textbf{62.77} & \hgreen{59.82}\\

    \cmidrule(lr){2-13}

    \multirow{-4}{*}{ViT-B/32 } & MaPLe & \textbf{88.37} & \textbf{66.43} & \textbf{18.47} & 42.03 & \textbf{37.77} & \textbf{60.00} & 80.10 & 64.43 & 91.33  & 62.50 & 61.14\\
    
    \rowcolor{tabhighlight} \cellcolor{white} & MaPLe + \method & 88.27 & 64.77 & 17.33 & \textbf{43.13} & 37.13 & 59.93 & \textbf{80.47} & \textbf{65.43} & \textbf{93.43}  & \textbf{62.70} & \hgreen{61.26}\\

    \cmidrule(lr){2-13}

    & PromptSRC & 88.13 & \textbf{65.33} & 17.17 & \textbf{43.90} & 40.50 & 60.17 & \textbf{81.23} & 65.40 & 92.37 & \textbf{63.63} & 61.78\\
        
    \rowcolor{tabhighlight} \cellcolor{white} & PromptSRC + \method  & \textbf{88.17} & 65.10 & \textbf{18.23} & 43.27 & \textbf{44.30} & \textbf{60.43} & 80.90 & \textbf{65.67} & \textbf{93.63} & 63.40 & \hgreen{62.31}\\

    \midrule

    ViT-H/14 & \demph{CLIP (teacher)} & \demph{94.80} & \demph{89.40} & \demph{63.20} & \demph{66.80} & \demph{63.30} & \demph{95.60} & \demph{93.60} & \demph{76.40} & \demph{97.90} & \demph{76.40} & \demph{81.74}\\
    
    \bottomrule
    \end{tabular}
}
\end{table*}
In \Cref{tab:cross-dataset} we present the results of the cross-dataset evaluation setting in which the prompt is trained on ImageNet and tested on ten different target datasets. For all datasets our \method-based variants consistently outperform the corresponding baselines and demonstrate superior generalization performance. The greater generalization capabilities of \method are evident even for fine-grained datasets like EuroSAT, on which the CoOp+\method achieves an 8\% improvement over the baseline CoOp when using ResNet-50 as the backbone. Although adding \method yields only minor improvement to the VPT and MaPLe prompt learning techniques, we emphasize that our VPT+\method and MaPLe+\method do not have access to ground-truth labels. Notably, PromptSRC+\method achieves the highest average performance.

\subsection{Results on Generalization to Unseen Classes}
\begin{table*}[t]
    \caption{\textbf{Generalization to Unseen Classes}. Comparison between the baselines and the proposed unsupervised \method variants (highlighted in cyan). In this setting the prompt is tested on classes never observed during training. Average performance improvements over the baselines are indicated in \hgreen{green}.}
    \label{tab:unseen-classes}
    \centering
    \resizebox{\textwidth}{!}{
    \begin{tabular}{c l cccccccccccc}
    \toprule
    &  & \multicolumn{11}{c}{Unseen Classes} \\ \cmidrule(lr){3-14}
    Backbone &
    Method & \rotatebox{80}{ImageNet} & \rotatebox{80}{OxfordPets} & \rotatebox{80}{Flowers102} & \rotatebox{80}{FGVCAircraft} & \rotatebox{80}{DTD} & \rotatebox{80}{EuroSAT} & \rotatebox{80}{StanfordCars} & \rotatebox{80}{Food101} & \rotatebox{80}{SUN397} & \rotatebox{80}{Caltech101} & \rotatebox{80}{UCF101} & \rotatebox{80}{\emph{\textit{Average}}} \\
    \midrule
    
     & \demph{CLIP (student)} & \demph{60.10} & \demph{93.70} & \demph{71.30} & \demph{24.80} & \demph{53.90} & \demph{43.70} & \demph{66.60} & \demph{82.20} & \demph{70.10} & \demph{90.50} & \demph{67.80} & \demph{65.88} \\
    \cmidrule(lr){2-14}

    & CoOp & 60.00 & 91.63 & 58.30 & \textbf{22.17} & \textbf{40.93} & 42.13 & \textbf{59.87} & 82.03 & 66.77 & 87.53 & 57.63 & 60.82 \\

    \rowcolor{tabhighlight} \cellcolor{white} & CoOp + KDPL & \textbf{60.53} & \textbf{94.97} & \textbf{63.40} & 21.27 & 38.70 & \textbf{62.07} & 56.63 & \textbf{83.07} & \textbf{69.83} & \textbf{88.75} & \textbf{62.63} & \hgreen{63.80} \\ 
 
    \cmidrule(lr){2-14}

    \multirow{-4}{*}{RN50} & CoCoOp & \textbf{63.07} & 94.93 & 67.47 & \textbf{24.87} & 45.40 & 35.10 & \textbf{64.73} & 84.50 & 72.87 & \textbf{90.73} & 64.67 & 64.39\\

    \rowcolor{tabhighlight} \cellcolor{white} & CoCoOp + KDPL  & 62.43 & \textbf{95.10} & \textbf{69.17} & 24.57 & \textbf{48.00} & \textbf{55.93} & 63.70 & \textbf{85.67} & \textbf{73.33} & 90.53 & \textbf{66.40} & \hgreen{66.80}\\ 
 
    \midrule

     & \demph{CLIP (student)} & \demph{64.00} & \demph{96.30} & \demph{72.30} & \demph{28.30} & \demph{53.90} & \demph{61.50} & \demph{69.80} & \demph{85.70} & \demph{73.10} & \demph{94.00} & \demph{71.60} & \demph{70.05} \\

    \cmidrule(lr){2-14}

     & CoOp & 64.53 & 93.37 & \textbf{59.77} & \textbf{23.03} & 45.57 & 49.13 & 61.73 & 85.50 & 69.37 & 92.13 & 65.60 & 64.52 \\
    
    \rowcolor{tabhighlight} \cellcolor{white} & CoOp + KDPL & \textbf{64.90} & \textbf{94.00} & 56.17 & 22.00 & \textbf{47.33} & \textbf{62.07} & \textbf{62.37} & \textbf{87.23} & \textbf{74.17} & \textbf{93.43} & \textbf{66.20} & \hgreen{66.35} \\

    \cmidrule(lr){2-14}

    \rowcolor{white} \cellcolor{white} & VPT & 64.60 & 94.83 & 66.03 & 29.70 & \textbf{50.17} & 50.03 & \textbf{69.90} & 85.83 & 75.13 & 92.57 & 69.47 & 68.02 \\
    
    \rowcolor{tabhighlight} \cellcolor{white} & VPT + KDPL & \textbf{65.20} & \textbf{95.20} & \textbf{67.87} & \textbf{30.40} & 47.53 & \textbf{54.97} & 69.80 & \textbf{86.33} & \textbf{75.17} & \textbf{92.80} & \textbf{71.07} & \hgreen{68.76} \\

    \cmidrule(lr){2-14}

    \multirow{-4}{*}{ViT-B/32 } & MaPLe & 66.70 & \textbf{96.87} & \textbf{69.77} & 20.43 & \textbf{53.50} & 66.70 & 68.03 & 87.47 & 76.57 & 92.27 & 69.40 & 69.79 \\
    
    \rowcolor{tabhighlight} \cellcolor{white} &  MaPLe + KDPL & \textbf{66.73} & \textbf{96.87} & 68.23 & \textbf{27.37} & 50.33 & \textbf{75.70} & \textbf{68.40} & \textbf{87.83} & \textbf{76.83} & \textbf{93.67} & \textbf{73.53} & \hgreen{71.41} \\

    \cmidrule(lr){2-14}

    & PromptSRC & 65.93 & 96.30 & \textbf{71.37} & 23.67 & \textbf{54.43} & \textbf{61.97} & \textbf{70.00} & 87.13 & 76.77 & \textbf{94.70} & 72.63 & 70.45\\
        
    \rowcolor{tabhighlight} \cellcolor{white} & PromptSRC + \method  & \textbf{66.33} & \textbf{96.47} & 68.83 & \textbf{28.43} & 50.87 & 74.17 & 68.50 & \textbf{87.63} & \textbf{77.33} & 94.47 & \textbf{74.73} & \hgreen{71.61}\\

    \midrule

    ViT-H/14 & \demph{CLIP (teacher)} & \demph{84.10} & \demph{99.20} & \demph{85.60} & \demph{64.30} & \demph{75.00} & \demph{82.60} & \demph{98.20} & \demph{96.30} & \demph{85.10} & \demph{97.30} & \demph{85.20} & \demph{86.63} \\

    \bottomrule
    \end{tabular}
}
\end{table*}
In \Cref{tab:unseen-classes} we give results for the unseen class generalization task. In these experiments each dataset is split into 50\% of the classes as a base for training few-shot adaptation, and the remaining 50\% as new classes on which zero-shot performance is evaluated. \method consistently outperforms the corresponding baseline methods for both backbones, demonstrating improvement in all scenarios for the majority of the datasets. On average, the performance improvement over the supervised baseline methods ranges from about 1\% for VPT to about 3\% for CoOp with the ResNet-50 backbone. See Section \ref{supp:unseen-classes} in the Supplementary Material for further analysis of base and unseen performance.

\subsection{Results on Class Agnostic Adaptation}

\begin{figure}[t]
    \begin{tabular}{cc}
    \begin{minipage}[t]{0.48\textwidth}
        \centering  
        \resizebox{1\textwidth}{!}{
    \begin{tabular}{l c cc}
        \toprule
        Method & Source & $\text{AVG}^1$ & $\text{AVG}^2$\\
        \midrule
        \demph{ResNet-50 (student)} & \demph{58.20} & \demph{40.63} & \demph{55.81}\\
        \rowcolor{tabhighlight}
        CoOp + \CAmethod & 61.83 & \hgreen{43.02} & \hgreen{57.69}\\
        \rowcolor{tabhighlight}
        CoCoOp + \CAmethod  & 61.43 & \hgreen{43.18 }& \hgreen{58.03}\\
        \midrule
        \rowcolor{white}
        \demph{ViT-B/32 (student)} & \demph{62.00} & \demph{47.78} & \demph{60.17} \\
        \rowcolor{tabhighlight}
        CoOp + \CAmethod & 64.73 & \hgreen{49.63} & \hgreen{60.74}\\
        \rowcolor{tabhighlight}
        VPT + \CAmethod & 63.73 & \hgreen{47.89} & \hred{59.80}\\
        \rowcolor{tabhighlight}
        MaPLe + \CAmethod & 65.23 & \hgreen{49.28} & \hgreen{61.23}\\

        \rowcolor{tabhighlight}
        PromptSRC + \CAmethod & 65.00 & \hgreen{50.37} & \hgreen{61.93}\\
      
        \bottomrule
    \end{tabular}
}

    \end{minipage}
    &
    \raisebox{-.35\height}{\includegraphics[width=0.45\textwidth]{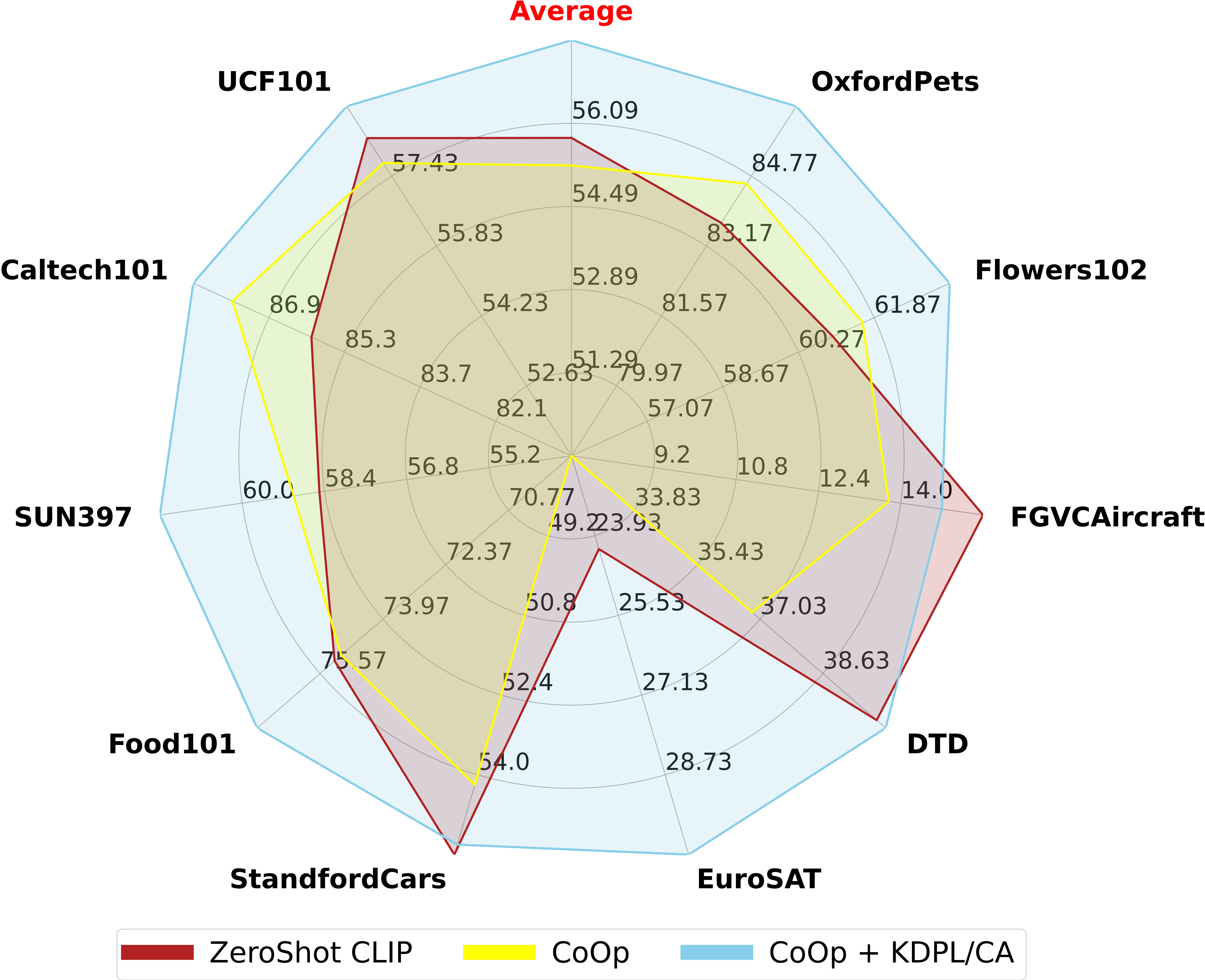}} \\
    & \\

    \textbf{(a)} Average class agnostic adaptation & \textbf{(b)} Per-dataset class agnostic adaptation
    \end{tabular}
    \caption{\textbf{Class agnostic adaptation}. \textbf{(a)} Comparison between the zero-shot baselines and our class agnostic \CAmethod variants (highlighted in cyan). The prompt is learned in an unsupervised and class agnostic setting on ImageNet and evaluated on the benchmark datasets for domain generalization ($\text{AVG}^1$) and cross-dataset ($\text{AVG}^2$) evaluations. Average performance improvements are indicated in \hgreen{green}, and deterioration in \hred{red}. \textbf{(b)} Per-dataset accuracy comparison between our unsupervised and class agnostic method (CoOp+\CAmethod), the supervised baseline CoOp, and the zero-shot student on the cross-dataset benchmark datasets.}
    \label{fig:class-agnostic}
\end{figure}

Figure~\ref{fig:class-agnostic}(a-b) summarizes the main results in the proposed Class Agnostic (CA) scenario in which even the training class names are unknown at training time. We report the accuracy on the ImageNet dataset (source), as well as the average accuracy in domain generalization and cross-dataset settings. Note that, even without knowing the class names, the performance on the source dataset steadily improves compared to the zero-shot CLIP model. Moreover, the prompts learned via the proposed unsupervised class agnostic knowledge distillation also exhibit improved average domain generalization ($\text{AVG}^1$) and cross-dataset capabilities ($\text{AVG}^2$).
Figure~\ref{fig:class-agnostic}(b) visually depicts how the ResNet-50-based CoOp+\CAmethod compares with supervised CoOp and zero-shot CLIP student performance in the cross-dataset transfer setting. Notably, \CAmethod outperforms both baselines despite being unsupervised and class agnostic during training.

\section{Conclusion}
\label{sec:discussion}

In this paper we proposed an approach to \extmethod (\method) which is easily integrated into existing supervised prompt learning methods. Our experiments show that for CoOp, CoCoOp, VPT, MaPLe and PromptSRC adding \method: (1) renders them \textit{label agnostic} by eliminating the need for ground-truth labels for few-shot adaptation; (2) can also render them \emph{class agnostic} in cases where no knowledge of training class labels is available; and (3) remarkably improves generalization to downstream tasks.

The additional computation cost incurred by distilling from a large VLM is a limitation of \method, though all encoder parameters are fixed and the predictions of the teacher can be precomputed, which helps mitigate the extra cost. This extra computation only amounts to about a 15\% increase for MaPLe.

Note that we did not tune any hyperparameters when training the \method variants. We use the same settings as in the original papers, which are likely suboptimal for distillation-based adaptation. With careful tuning, there is potential for improvement. Our experiments indicate that distillation-learned prompts are more transferable, and we think it would be interesting to see if this idea can generalize to very different downstream tasks and scale to even bigger teacher models or -- more interestingly -- to even smaller student models.

\clearpage
\section*{Acknowledgements}
This work was supported by funding from the European Commission Horizon 2020 grant \#951911
(AI4Media).

\bibliographystyle{splncs04}
\bibliography{egbib}
\newpage

\title{Supplementary Material for: \\ Improving Zero-shot Generalization of Learned Prompts via Unsupervised Knowledge Distillation} 
\titlerunning{Improving Learned Prompts via Unsupervised Knowledge Distillation}
\def\thefootnote{*}\footnotetext{These authors contributed equally to this work.}
\def\thefootnote{\arabic{footnote}}
\author{$^*$Marco Mistretta\inst{1}\orcidlink{0009-0006-6630-6477} \and
$^*$Alberto Baldrati\inst{1,2}\orcidlink{0000-0002-5012-5800} \and \\ 
Marco Bertini\inst{1}\orcidlink{0000-0002-1364-218X} \and
Andrew D. Bagdanov\inst{1}\orcidlink{0000-0001-6408-7043}
}
\authorrunning{M.~Mistretta \emph{et al.}}
\institute{University of Florence - Media Integration and Communication Center (MICC) \and
University of Pisa \\
Florence, Italy - Pisa, Italy \\
\email{\{name.surname\}@unifi.it}}
\maketitle

\section*{Overview}
This document contains Supplementary Material that provides additional details and further experimental analysis. The contents are organized as follows:
\begin{itemize} 
    \item \textbf{\ref{supp:impl-details} Additional Implementation Details}: This section provides additional implementation details, including hyperparameters and configuration settings for reproducing all reported results and further details about the datasets used.

    \item \textbf{\ref{supp:ablations} Ablation Study}
    \begin{itemize}
    \item \ref{supp:pseudolabeling} \textbf{Pseudolabelling versus \method}: This subsection compares \method performances with a famous pseudolabeling prompt learning strategy demonstrating the superiority of \method over the baseline approach.

    \item \ref{supp:sampling} \textbf{Sampling Strategies for Class Agnostic Adaptation}: This part evaluates different sampling strategies for class agnostic adaptation and their performance varying the number of sampled class names.
 
    \item \ref{supp:symmetric} \textbf{Symmetric versus Asymmetric KL}: This section explores the impact of using symmetric versus asymmetric KL-divergence in the loss function explaining why is more effective for class agnostic scenarios.

    \item \ref{supp:teacher} \textbf{Is a smaller Teacher enough?} Here, we investigate whether smaller teachers model can achieve competitive performance compared to the larger ViT-H-14 teacher model used in the main paper. 

    \end{itemize}
    
    \item \textbf{\ref{supp:additional} Additional Results}
    
    \begin{itemize}
        \item \ref{supp:unseen-classes} \textbf{Generalization to Unseen Classes}: Here we present additional experimental results for the generalization to unseen classes within the same dataset scenario, including the downstream performance of few-shot base as well as the harmonic mean of base and unseen classes.

        \item \ref{supp:class-agnostic} \textbf{Class Agnostic Scenario}: This section provides additional results for the class agnostic setting, our proposed new setting in which we know neither the labels nor the names of the training classes.
    \end{itemize}
    
\end{itemize}

\asection{Additional Implementation Details}
\label{supp:impl-details}

% Set the figure counter to 3
\setcounter{figure}{3}
% Set the table counter to 3
\setcounter{table}{3}
\begin{table}[t]
\centering
\begin{minipage}{.49\linewidth}
\centering
\caption{Configuration of hyperparameters for the re-computed baselines. Note that CoOp and CoCoOp, being textual prompt learning methods, do not learn visual tokens.}
\label{tab:bas-hyp}
    \resizebox{\textwidth}{!}{
        \begin{tabular}{@{}lccccc@{}}
            \toprule
            Dataset      & CoOp & CoCoOp & VPT  & MaPLe & PromptSRC \\ \midrule
            Batch Size     & 32   & 1 & 4  & 4 & 4 \\
            Optimizer   & SGD & SGD & SGD & SGD & SGD \\
            LR   & 0.02      & 0.02 & 0.0025   & 0.0035 & 0.0025 \\
            Epochs & 50     & 10 & 5 & 5 & 20 \\
            Text-tokens   & 4     & 4 & 4 & 4 & 4 \\
            Visual-tokens      & ---     & --- & 8 & 2 & 4 \\
            Vision-depth & ---     & --- & 12 & 9 & 9 \\
            \bottomrule
        \end{tabular}
    }
    \vspace{0.39in}
\end{minipage}
\hfill
\begin{minipage}{.49\linewidth}
\centering
\caption{Datasets statistics.}
\label{tab:dataset-stat}
    \resizebox{0.9\textwidth}{!}{
        \begin{tabular}{@{}lcccc@{}}
            \toprule
            Dataset      & Classes & Train & Val  & Test \\ \midrule
            ImageNet~\cite{deng2009imagenet}     & 1,000   & 1.28M & N/A  & 50,000 \\
            Caltech101~\cite{fei2004learning}   & 100     & 4,128 & 1,649 & 2,465 \\
            OxfordPets~\cite{parkhi2012cats}   & 37      & 2,944 & 736   & 3,669 \\
            StanfordCars~\cite{krause20133d} & 196     & 6,509 & 1,635 & 8,041 \\
            Flowers102~\cite{nilsback2008automated}   & 102     & 4,093 & 1,633 & 2,463 \\
            Food101~\cite{bossard2014food}      & 101     & 50,500 & 20,300 & 30,300 \\
            FGVCAircraft~\cite{maji2013fine} & 100     & 3,334 & 3,333 & 3,333 \\
            SUN397~\cite{xiao2010sun}       & 397     & 15,880 & 3,970 & 19,850 \\
            DTD~\cite{cimpoi2014describing}          & 47      & 2,820 & 1,128 & 1,692 \\
            EuroSAT~\cite{helber2019eurosat}      & 10      & 13,500 & 5,400 & 8,100 \\
            UCF101~\cite{soomro2012ucf101}       & 101     & 7,639 & 1,898 & 3,783 \\
            \midrule
            ImageNetV2~\cite{recht2019imagenet}   & 1,000   & N/A   & N/A   & 10,000 \\
            ImageNet-S~\cite{wang2019learning} & 1,000 & N/A   & N/A   & 50,889 \\
            ImageNet-A~\cite{hendrycks2021natural}   & 200     & N/A   & N/A   & 7,500 \\
            ImageNet-R~\cite{hendrycks2021many}   & 200     & N/A   & N/A   & 30,000 \\ \bottomrule
        \end{tabular}
    }

\end{minipage}

\end{table}

In this section we provide further details on hyperparameters of the proposed approaches presented in the main paper and additional details on the benchmark datasets evaluated.

\minisection{Hyperparameter settings} For all evaluated scenarios we recalculated all baseline results using the parameter and hyperparameter settings reported in the original publications~\cite{zhou2022learning, zhou2022conditional, jia2022visual, khattak2023maple, khattak2023self}. 

In \Cref{tab:bas-hyp} we summarize all configuration settings. Notably, all the evaluated baselines use a cosine scheduler with one warm-up epoch with a constant learning rate of 1e-5, and all authors suggest initializing the textual context with ``a photo of a'' as a better starting point. Note that our variants augmented with Knowledge Distillation Prompt Learning (\method) were trained using exactly the same parameters and configuration settings as the baseline methods. Therefore, it is reasonable to expect that further hyperparameter tuning could potentially improve our performance even further.

In addition, all the baselines -- and so also our augmented versions -- adhere to the same preprocessing pipeline in which input images are resized to $224 \times 224$ pixels using bicubic interpolation. Data augmentations including random resized crop, random flip, and normalization are used to ensure robustness.

To ensure consistency with baselines and to facilitate easy comparison and evaluation,
our implementation is built upon the Dassl framework\footnote{\url{https://github.com/KaiyangZhou/Dassl.pytorch}}, which serves as the foundation for all these methods. We use ResNet-50 and ViT-B/32 as the backbones for all experiments. These backbones are the original publicly available models released by OpenAI.\footnote{\url{https://github.com/openai/CLIP}}
As mentioned in the main paper, the code is publicly available at \small{\href{https://github.com/miccunifi/KDPL}{\url{https://github.com/miccunifi/KDPL}}}.

\minisection{Dataset Details.} The detailed statistics of the 11 datasets used in the Cross-dataset transfer setting, as well as the four variants of ImageNet used in the Cross-domain generalization setting, are given in \Cref{tab:dataset-stat}. Note that according to the authors of CoOp, for Caltech101, the ``BACKGROUND Google'' and ``Faces easy'' classes are discarded, and for the video dataset UCF101 the middle frame of each video is used as input to the image encoder~\cite{zhou2022learning}.

\asection{Ablation Study}
\label{supp:ablations}

\begin{table}[t]
  \caption{\textbf{Ablations on labeling, sampling, and KL-divergence}. \textbf{(a)} Comparison of KDPL and a pseudolabeling strategy adapted from UPL~\cite{huang2022unsupervised}. \textbf{(b)} Ablation on a number of classes sampled for class agnostic adaptation. \textbf{(c)} Comparison of forward, reverse, and symmetric KL-divergence used in the loss (see Eq. 2 of the main paper).}
  \label{tab:ablations}
  
  \centering
  \begin{tabular}{ccc}
    \begin{minipage}[t]{.32\textwidth}
      \centering
      \resizebox{!}{0.95cm}{
    \begin{tabular}{l c ccc}
        \toprule
        & Source & $\text{AVG}^1$ & $\text{AVG}^2$\\
        \toprule
        \demph{CLIP (student)} &  \demph{58.20} & \demph{40.63} & \demph{55.81} \\ 
        
        \midrule
        
        \rowcolor{white} CoOp + UPL* & 62.37 & 42.72 & 56.24\\
        
        \rowcolor{tabhighlight} CoOp + \method & \textbf{62.73} & \textbf{42.90} & \textbf{57.59}\\
        
        \midrule
        
        \rowcolor{white} CoOp + CA-UPL* & 60.13 & 42.48 & 55.73\\
        
        \rowcolor{tabhighlight}  CoOp + \CAmethod & \textbf{61.83} & \textbf{43.02} & \textbf{57.69}\\
        
        \bottomrule
    \end{tabular}
}
    \end{minipage}
    &
    \begin{minipage}[t]{.32\textwidth}
      \centering
      \resizebox{!}{0.95cm}{
    \begin{tabular}{l c cc}
        \toprule
        & Source & $\text{AVG}^1$ & $\text{AVG}^2$\\
        \toprule
        POMP* & 60.80 & 42.37 & 57.35 \\
        \midrule
        \rowcolor{white} \CAmethod ($K$=100) & 61.50 & 42.45 & 57.06\\
        \rowcolor{white} \CAmethod ($K$=500) & 61.73 & 42.68 & 57.38\\
        
        \rowcolor{tabhighlight} \CAmethod ($K$=1K) & \textbf{61.83} & 43.02 & \textbf{57.69}\\
        
        \rowcolor{white} \CAmethod ($K$=2K) & 61.60 & \textbf{43.07} & 57.51\\
        \bottomrule
 
    \end{tabular}
}

    \end{minipage}
    &
    \begin{minipage}[t]{.32\textwidth}
      \centering
      \resizebox{!}{0.95cm}{
    \begin{tabular}{l c ccc}
        \toprule
        & Source & $\text{AVG}^1$ & $\text{AVG}^1$\\
        \toprule
        
        \rowcolor{white} CoOp + \method F & \textbf{63.13} & \textbf{43.16} & 57.33\\
        
        \rowcolor{white} CoOp + \method R & 62.47 & 42.44 & 57.24\\
        
        \rowcolor{tabhighlight} CoOp + \method & 62.73 & 42.90 & \textbf{57.59}\\
        
        \midrule
        
        \rowcolor{white} CoOp + \CAmethod F & 60.77 & 42.68 & 56.39\\
        
        \rowcolor{white} CoOp + \CAmethod R & 61.37 & 42.71 & 57.12\\
        
        \rowcolor{tabhighlight} CoOp + \CAmethod & \textbf{61.83} & \textbf{43.02} & \textbf{57.69}\\
        
        \bottomrule
    \end{tabular}
}
     \end{minipage}

    \\
    
    \scriptsize \textbf{(a)} KDPL vs. pseudolabeling
    &
    \scriptsize \textbf{(b)} Sampling strategy
    &
    \scriptsize \textbf{(c)} Symmetric/asymmetric KL
    
  \end{tabular}

  \setlength{\abovecaptionskip}{0pt}
  \setlength{\belowcaptionskip}{0pt}
\end{table}

We performed a range of ablation studies to evaluate all aspects of \method. All ablations reported here are averages over three independent runs using CoOp as the baseline approach and ResNet-50 as the student backbone. The prompt is learned on ImageNet and evaluated on the benchmark datasets used in the domain generalization ($\text{AVG}^1$) and cross-dataset ($\text{AVG}^2$) evaluations.

\asubsection{Pseudolabeling versus \method}
\label{supp:pseudolabeling}
We compare \method and \CAmethod with a pseudolabeling strategy for exploiting unlabeled samples. Similar to the approach outlined in UPL~\cite{huang2022unsupervised}, we generate teacher-derived labels for the training samples. Instead of selecting the top-K confident pseudolabels over the entire training set, for a fair comparison we pseudolabel the fixed, few shot samples in each run. We call this baseline UPL*. 
Table~\ref{tab:ablations}(a) shows 
that \method consistently outperforms UPL* on all the evaluated scenarios.

\asubsection{Sampling Strategies for Class Agnostic Adaptation}
\label{supp:sampling}
An important detail in the class agnostic adaptation scenario is how class names are selected for computing the loss. Existing supervised approaches, such as POMP~\cite{ren2024prompt},  to handling large numbers of classes suggest simply using the ground truth labels of batch samples and supplementing them with randomly selected ones to reach a predetermined number $K$. We adapted this technique and compare it with ours in Table~\ref{tab:ablations}(b). These results show that \method outperforms the random selection strategy denoted as POMP*. Notably, the optimal number of classes for domain generalization and cross-dataset adaptation is 1000, suggesting that the optimum closely aligns with the actual number of classes in the training dataset.

\asubsection{Symmetric versus Asymmetric KL} 
\label{supp:symmetric}
In Eq. 2 of the main paper we propose to use the symmetric KL-divergence, which is a sum of the forward (from teacher to student) and reverse (from student to teacher) KL-divergences. 
An asymmetric forward loss may yield benefits when the target distribution is similar to the source distribution (see Table~\ref{tab:ablations}(c)). 
Conversely, a reverse loss may prove better in the opposite scenario. 
If the Teacher outputs probability zero for a class on a training sample, the \textit{forward KL-divergence} is zero for that sample even if the Student outputs a very high probability (see Eq.~(3), main paper). Adding the \textit{reverse KL-divergence} prevents the distillation loss from going to zero in such cases (Eq.~(2), main paper).
Indeed, in the class agnostic scenario the optimal strategy is the symmetric KL divergence. This enables learning from both in- and out-of-domain training samples, as demonstrated by the results in Figure~3 of the main paper.

\asubsection{Is a smaller Teacher enough?}
\label{supp:teacher}
All the results presented so far involve \mbox{ViT-H-14} as the teacher.
To evaluate if a smaller teacher would be enough we ran additional experiments with \mbox{ViT-L/14} and \mbox{ViT-B/16} as Teacher and ResNet-50 as Student (see Table~\ref{tab:teacher}). Larger Teachers yield better results, but even more modest ones still maintain excellent performance on domain generalization and cross-dataset transfer. Interestingly, smaller Teachers work \emph{better} on generalization to unseen classes, which we believe is due to the weaker supervision causing less adaptation and thus maintaining better zero-shot transfer.
\begin{table}[t]

\centering
\caption{Average accuracy comparison on the three evaluated scenarios between the supervised baseline CoOp and our unsupervised approach (CoOp + \method) using different and smaller teachers.}
\label{tab:teacher}

\resizebox{0.8\linewidth}{!}{
\begin{tabular}{l c ccc}

    \toprule
        &  & \multicolumn{3}{c}{Target downstream \textit{Average}} \\
    \cmidrule(lr){3-5} 
    
    Method & Source  & \shortstack{Domain\\Generalization} & \shortstack{Cross-Dataset\\Transfer} & \shortstack{Generalization to\\Unseen Classes} \\
    \midrule
    
    CoOp & 62.40 & 42.05 & 55.28 & 60.82\\

    \midrule
    
    \rowcolor{tabhighlight}
    CoOp + \method (B/16)  & 60.73 & 42.27 & 56.52 & \textbf{67.57}\\
    
    \rowcolor{tabhighlight}
    CoOp + \method (L/14)  & 61.43 & 42.61 & 57.10 & 66.90\\
    
    \rowcolor{tabhighlight}
    CoOp + \method (H/14)  & \textbf{62.73} & \textbf{42.90} & \textbf{57.59} & 63.80\\
    \bottomrule
\end{tabular}
}

\end{table}

\asection{Additional Results}
\label{supp:additional}
\setlength{\tabcolsep}{5pt}
\begin{table*}
  \centering
  \caption{
  \textbf{Base-to-new generalization.} Prompts are learned from the base classes (16-shots) and evaluated on the same classes on the test set (Base), zero-shot on the unseen classes of the test set (New), and in terms of harmonic mean (H) between the base and the new accuracy. Average \textit{New} performance improvements over the baselines are indicated in \hgreen{green}. Here we see how \method generalizes better to the new classes, limiting deterioration in performance caused by the unbalanced few-shot training on half the classes of the source domain. Remarkably, we are competitive even on the base classes without requiring training labels.
  }
  \resizebox{1.0\textwidth}{!}{
   \label{tab:base-to-new}
    \begin{tabular}{llrrrrrrrrrrrr}
    \toprule
    \multicolumn{2}{c}{\multirow{2}[2]{*}{}} & \multicolumn{3}{c}{\textbf{Average}} & \multicolumn{3}{c}{ImageNet} & \multicolumn{3}{c}{OxfordPets} & \multicolumn{3}{c}{Flowers102} \\    \cmidrule(lr){3-5} \cmidrule(lr){6-8} \cmidrule(lr){9-11} \cmidrule(lr){12-14}
    
    \multicolumn{2}{c}{} & \multicolumn{1}{c}{Base} & \multicolumn{1}{c}{\textbf{New}} & \multicolumn{1}{c}{H} & \multicolumn{1}{c}{Base} & \multicolumn{1}{c}{New} & \multicolumn{1}{c}{H} & \multicolumn{1}{c}{Base} & \multicolumn{1}{c}{New} & \multicolumn{1}{c}{H} & \multicolumn{1}{c}{Base} & \multicolumn{1}{c}{New} & \multicolumn{1}{c}{H} \\
    \midrule

     & \demph{CLIP (student)} & \demph{61.61} & \demph{65.88} & \demph{63.67} & \demph{64.40} & \demph{60.10} & \demph{58.20} & \demph{85.80} & \demph{93.70} & \demph{83.70} & \demph{63.80} & \demph{71.30} & \demph{61.00}\\

    \cmidrule(lr){2-14}
    
    & CoOp & \textbf{77.13} & 60.82 & 68.01 & \textbf{68.40} & 60.00 & 60.30 & 90.80 & 91.63 & 85.03 & \textbf{95.17} & 58.30 & 67.03\\
    
    \rowcolor{tabhighlight} \cellcolor{white} &  CoOp + \method & 73.66 & \hgreen{63.80} & \textbf{68.38} & 68.37 & \textbf{60.53} & \textbf{60.63} & \textbf{93.13} & \textbf{94.97} & \textbf{88.63} & 93.93 & \textbf{63.40} & \textbf{69.77}\\

    \cmidrule(lr){2-14}
    
    \multirow{-4}{*}{RN50} & CoCoOp & \textbf{75.47} & 64.39 & \textbf{69.49} & \textbf{68.30} & \textbf{63.07} & \textbf{61.90} & 92.13 & 94.93 & 87.73 & \textbf{91.10} & 67.47 & 68.37\\
    
    \rowcolor{tabhighlight} \cellcolor{white} & CoCoOp + \method & 71.69 & \hgreen{65.90} & 68.67 & 68.13 & 62.87 & 61.63 & \textbf{93.40} & \textbf{95.33} & \textbf{89.07} & 87.77 & \textbf{70.63} & \textbf{70.43}\\ 
    
    \midrule
    
    & \demph{CLIP (student)} & \demph{64.81} & \demph{70.05} & \demph{67.33} & \demph{67.40} & \demph{64.00} & \demph{62.00} & \demph{86.90} & \demph{96.30} & \demph{85.00} & \demph{68.70} & \demph{72.30} & \demph{64.30}\\

    \cmidrule(lr){2-14}

    & CoOp & \textbf{79.02} & 64.52 & \textbf{71.04} & \textbf{70.90} & 64.53 & \textbf{63.90} & 91.90 & 93.37 & 84.77 & \textbf{95.13} & \textbf{59.77} & \textbf{67.30}\\
    
    \rowcolor{tabhighlight} \cellcolor{white} & CoOp + \method & 75.20 & \hgreen{66.35} & 70.50 & 70.57 & \textbf{64.90} & 63.83 & \textbf{93.67} & \textbf{94.00} & \textbf{87.57} & 93.47 & 56.17 & 65.23\\

    \cmidrule(lr){2-14}
    
    & VPT & \textbf{75.15} & 68.02 & \textbf{71.41} & 70.30 & 64.60 & 63.85 & \textbf{94.03} & 94.83 & \textbf{85.87} & \textbf{88.10} & 66.03 & \textbf{67.73}\\
    
    \rowcolor{tabhighlight} \cellcolor{white} & VPT + \method & 72.12 & \hgreen{68.76} & 70.40 & \textbf{70.70} & \textbf{65.20} & \textbf{64.30} & 93.80 & \textbf{95.20} & 85.53 & 82.73 & \textbf{67.87} & 67.30\\

    \cmidrule(lr){2-14}
    
    \multirow{-4}{*}{ViT-B/32 } & MaPLe & \textbf{78.26} & 69.79 & \textbf{73.79} & 71.40 & 66.70 & 65.40 & 93.07 & \textbf{96.87} & 90.03 & \textbf{93.37} & \textbf{69.77} & \textbf{71.83}\\
    
    \rowcolor{tabhighlight} \cellcolor{white} & MaPLe + \method & 74.74 & \hgreen{71.41} & 73.03 & \textbf{71.60} & \textbf{66.73} & \textbf{65.53} & \textbf{93.90} & \textbf{96.87} & \textbf{90.43} & 89.73 & 68.23 & 69.70\\

    \cmidrule(lr){2-14}
    
    & PromptSRC & \textbf{81.17} & 70.45 & \textbf{75.43} & 72.50 & 65.93 & 65.63 & 93.40 & 96.30 & 89.30 & \textbf{96.20} & \textbf{71.37} & \textbf{74.90}\\
    
    \rowcolor{tabhighlight} \cellcolor{white} & PromptSRC + \method & 77.11 & \hgreen{71.61} & 74.26 & \textbf{72.70} & \textbf{66.33} & \textbf{65.93} & \textbf{94.37} & \textbf{96.47} & \textbf{90.17} & 94.93 & 68.83 & 73.10\\
    
    \midrule
    ViT-H/14 & \demph{CLIP (teacher)} & \demph{86.11} & \demph{86.63} & \demph{86.37} & \demph{86.50} & \demph{84.10} & \demph{82.80} & \demph{94.40} & \demph{99.20} & \demph{94.80} & \demph{96.60} & \demph{85.60} & \demph{89.40}\\
    \bottomrule
    \end{tabular}
  }
    \vspace{0.2cm} 
    \vfill
   \resizebox{1.0\textwidth}{!}{
    \begin{tabular}{llrrrrrrrrrrrr}
    \toprule
    \multicolumn{2}{c}{\multirow{2}[2]{*}{}} & \multicolumn{3}{c}{FGVCAircraft} & \multicolumn{3}{c}{DTD} & \multicolumn{3}{c}{EuroSAT} & \multicolumn{3}{c}{StanfordCars} \\
    \cmidrule(lr){3-5} \cmidrule(lr){6-8} \cmidrule(lr){9-11} \cmidrule(lr){12-14}\\
    \multicolumn{2}{c}{} & \multicolumn{1}{c}{Base} & \multicolumn{1}{c}{New} & \multicolumn{1}{c}{H} & \multicolumn{1}{c}{Base} & \multicolumn{1}{c}{New} & \multicolumn{1}{c}{H} & \multicolumn{1}{c}{Base} & \multicolumn{1}{c}{New} & \multicolumn{1}{c}{H} & \multicolumn{1}{c}{Base} & \multicolumn{1}{c}{New} & \multicolumn{1}{c}{H} \\
   
    \midrule

    & \demph{CLIP (student)} & \demph{17.20} & \demph{24.80} & \demph{15.50} & \demph{49.00} & \demph{53.90} & \demph{40.00} & \demph{39.20} & \demph{43.70} & \demph{24.20} & \demph{55.40} & \demph{66.60} & \demph{55.60}\\

    \cmidrule(lr){2-14}

    & CoOp & \textbf{27.93} & \textbf{22.17} & \textbf{18.90} & \textbf{75.03} & \textbf{40.93} & \textbf{47.13} & \textbf{89.30} & 42.13 & 46.07 & 67.77 & \textbf{59.87} & \textbf{58.30}\\
    
    \rowcolor{tabhighlight} \cellcolor{white} & CoOp + \method & 24.33 & 21.27 & 17.03 & 66.73 & 38.70 & 42.70 & 71.17 & \textbf{62.07} & \textbf{47.23} & \textbf{67.93} & 56.63 & 56.70\\
    
    \cmidrule(lr){2-14}

    \multirow{-4}{*}{RN50} & CoCoOp & \textbf{24.27} & \textbf{24.87} & \textbf{18.30} & \textbf{72.37} & 45.40 & \textbf{45.73} & \textbf{87.73} & 35.10 & \textbf{42.00} & \textbf{63.43} & \textbf{64.73} & \textbf{58.87}\\

    \rowcolor{tabhighlight} \cellcolor{white} & CoCoOp + \method & 21.77 & 22.67 & 16.93 & 62.70 & \textbf{49.07} & 44.07 & 69.73 & \textbf{44.27} & 37.67 & 62.40 & 63.67 & 57.53\\ 
    
    \midrule
    
     & \demph{CLIP student} & \demph{20.30} & \demph{28.30} & \demph{18.20} & \demph{53.20} & \demph{53.90} & \demph{42.80} & \demph{43.40} & \demph{61.50} & \demph{38.10} & \demph{61.00} & \demph{69.80} & \demph{60.40}\\

    \cmidrule(lr){2-14}

    & CoOp & \textbf{30.80} & \textbf{23.03} & \textbf{20.27} & \textbf{77.17} & 45.57 & \textbf{49.50} & \textbf{88.37} & 49.13 & \textbf{51.23} & \textbf{71.13} & 61.73 & \textbf{61.17}\\

    \rowcolor{tabhighlight} \cellcolor{white} &  CoOp + \method &  26.67 & 22.00 & 18.23 & 67.13 & \textbf{47.33} & 44.10 & 71.40 & \textbf{62.07} & 46.90 & 69.37 & \textbf{62.37} & 60.47\\

    \cmidrule(lr){2-14}

    & VPT & \textbf{25.23} & 29.70 & \textbf{20.77} & \textbf{72.93} & \textbf{50.17} & \textbf{46.27} & \textbf{76.27} & 50.03 & \textbf{43.87} & \textbf{64.07} & \textbf{69.90} & \textbf{61.27}\\

    \rowcolor{tabhighlight} \cellcolor{white} &  VPT + \method & 24.10 & \textbf{30.40} & 20.03 & 63.40 & 47.53 & 42.07 & 65.43 & \textbf{54.97} & 42.40 & 63.77 & 69.80 & 61.23\\
    
    \cmidrule(lr){2-14}

    \multirow{-4}{*}{ViT-B/32 } & MaPLe & 23.57 & 20.43 & 16.50 & \textbf{77.27} & \textbf{53.50} & \textbf{53.90} & \textbf{91.17} & 66.70 & \textbf{63.93} & \textbf{67.50} & 68.03 & 62.50\\

    \rowcolor{tabhighlight} \cellcolor{white} &  MaPLe + \method & \textbf{25.43} & \textbf{27.37} & \textbf{19.93} & 66.27 & 50.33 & 46.53 & 74.03 & \textbf{75.70} & 62.30 & 66.97 & \textbf{68.40} & \textbf{62.40}\\

    \cmidrule(lr){2-14}
    
    & PromptSRC & \textbf{33.77} & 23.67 & 22.50 & \textbf{80.57} & \textbf{54.43} & \textbf{54.67} & \textbf{93.77} & 61.97 & \textbf{61.30} & \textbf{72.90} & \textbf{70.00} & \textbf{66.37}\\
    
    \rowcolor{tabhighlight} \cellcolor{white} & PromptSRC + \method & 30.37 & \textbf{28.43} & \textbf{23.20} & 71.93 & 50.87 & 47.97 & 73.93 & \textbf{74.17} & 58.53 & 72.53 & 68.50 & 65.43\\

    \cmidrule(lr){2-14}

    ViT-H/14 & \demph{CLIP (teacher)} & \demph{71.40} & \demph{64.30} & \demph{63.20} & \demph{78.20} & \demph{75.00} & \demph{66.80} & \demph{70.70} & \demph{82.60} & \demph{63.30} & \demph{94.60} & \demph{98.20} & \demph{95.60}\\
    \bottomrule
    \end{tabular}
  }
    \vspace{0.2cm} \vfill
    \resizebox{1.0\textwidth}{!}{
    \begin{tabular}{llrrrrrrrrrrrr}
    \toprule
    \multicolumn{2}{c}{\multirow{2}[2]{*}{}} & \multicolumn{3}{c}{Food101} & \multicolumn{3}{c}{SUN397} & \multicolumn{3}{c}{Caltech101} & \multicolumn{3}{c}{UCF101} \\
    \cmidrule(lr){3-5} \cmidrule(lr){6-8} \cmidrule(lr){9-11} \cmidrule(lr){12-14}\\
    \multicolumn{2}{c}{} & \multicolumn{1}{c}{Base} & \multicolumn{1}{c}{New} & \multicolumn{1}{c}{H} & \multicolumn{1}{c}{Base} & \multicolumn{1}{c}{New} & \multicolumn{1}{c}{H} & \multicolumn{1}{c}{Base} & \multicolumn{1}{c}{New} & \multicolumn{1}{c}{H} & \multicolumn{1}{c}{Base} & \multicolumn{1}{c}{New} & \multicolumn{1}{c}{H} \\
    \midrule
  
     & \demph{CLIP (student)} & \demph{81.60} & \demph{82.20} & \demph{75.20} & \demph{66.50} & \demph{70.10} & \demph{58.50} & \demph{91.10} & \demph{90.50} & \demph{86.00} & \demph{63.70} & \demph{67.80} & \demph{58.30}\\

    \cmidrule(lr){2-14}

    & CoOp & 82.63 & 82.03 & 75.37 & \textbf{76.37} & 66.77 & 61.50 & \textbf{95.50} & 87.53 & 88.53 & \textbf{79.50} & 57.63 & \textbf{61.53}\\
    
    \rowcolor{tabhighlight} \cellcolor{white} & CoOp + \method & \textbf{83.73} & \textbf{83.07} & \textbf{76.77} & 73.43 & \textbf{69.83} & \textbf{61.93} & 95.10 & \textbf{88.75} & \textbf{88.90} & 72.43 & \textbf{62.63} & 59.97\\

    \cmidrule(lr){2-14}

    \multirow{-4}{*}{RN50} & CoCoOp & 84.43 & 84.50 & 78.00 & \textbf{74.53} & 72.87 & \textbf{63.93} & \textbf{94.97} & 90.73 & 89.33 & \textbf{76.90} & 64.67 & \textbf{63.77}\\
    
    \rowcolor{tabhighlight} \cellcolor{white} & CoCoOp + \method & \textbf{84.63} & \textbf{85.90} & \textbf{79.00} & 71.83 & \textbf{73.23} & 62.80 & 94.87 & \textbf{91.00} & \textbf{89.37} & 71.33 & \textbf{66.23} & 61.70\\ 
    \midrule
    
     & \demph{CLIP student} & \demph{84.50} & \demph{85.70} & \demph{79.10} & \demph{69.80} & \demph{73.10} & \demph{62.00} & \demph{93.70} & \demph{94.00} & \demph{91.10} & \demph{64.00} & \demph{71.60} & \demph{60.70}\\

    \cmidrule(lr){2-14}
    
    & CoOp & 85.17 & 85.50 & 79.27 & \textbf{79.17} & 69.37 & 64.93 & \textbf{97.40} & 92.13 & 92.37 & \textbf{82.13} & 65.60 & \textbf{66.93}\\
    
    \rowcolor{tabhighlight} \cellcolor{white} &  CoOp + \method & \textbf{86.40} & \textbf{87.23} & \textbf{81.10} & 75.87 & \textbf{74.17} & \textbf{65.70} & 96.47 & \textbf{93.43} & \textbf{92.57} & 76.17 & \textbf{66.20} & 64.20\\

    \cmidrule(lr){2-14}

    & VPT & 85.47 & 85.83 & 79.60 & \textbf{75.70} & 75.13 & \textbf{65.47} & \textbf{97.10} & 92.57 & \textbf{92.80} & \textbf{77.43} & 69.47 & \textbf{65.10}\\
    
    \rowcolor{tabhighlight} \cellcolor{white} & VPT + \method & \textbf{85.63} & \textbf{86.33} & \textbf{79.97} & 74.07 & \textbf{75.17} & 65.07 & 96.33 & \textbf{92.80} & 92.53 & 73.33 & \textbf{71.07} & 63.70\\

    \cmidrule(lr){2-14}

    \multirow{-4}{*}{ViT-B/32 } & MaPLe & 86.40 & 87.47 & 81.27 & \textbf{78.90} & 76.57 & \textbf{68.60} & \textbf{97.03} & 92.27 & 93.00 & \textbf{81.23} & 69.40 & \textbf{68.17}\\
    
    \rowcolor{tabhighlight} \cellcolor{white} & MaPLe + \method & \textbf{86.50} & \textbf{87.83} & \textbf{81.73} & 75.57 & \textbf{76.83} & 67.07 & 97.00 & \textbf{93.67} & 93.33 & 75.10 & \textbf{73.53} & 67.07\\

    \cmidrule(lr){2-14}
    
    & PromptSRC & 86.33 & 87.13 & 80.93 & \textbf{80.80} & 76.77 & \textbf{69.90} & \textbf{97.53} & \textbf{94.70} & \textbf{94.30} & \textbf{85.07} & 72.63 & \textbf{72.67}\\
    
    \rowcolor{tabhighlight} \cellcolor{white} & PromptSRC + \method & \textbf{86.53} & \textbf{87.63} & \textbf{81.27} & 76.53 & \textbf{77.33} & 68.00 & 96.90 & 94.47 & 93.93 & 77.50 & \textbf{74.73} & 68.53\\

    \cmidrule(lr){2-14}

    ViT-H/14 & \demph{CLIP (teacher)} & \demph{95.50} & \demph{96.30} & \demph{93.60} & \demph{82.00} & \demph{85.10} & \demph{76.40} & \demph{99.20} & \demph{97.30} & \demph{97.90} & \demph{78.10} & \demph{85.20} & \demph{76.40}\\
    \bottomrule
    \end{tabular}
  }
  \label{tab:b2n}
\end{table*}

In the following sections we report on additional experiments validating our proposed approach.

\asubsection{Generalization to Unseen Classes}
\label{supp:unseen-classes}
In the main paper, due to brevity and to maintain focus on the core aspects of our method, we reported only the performance metrics on unseen classes. Here we provide the performance metrics on base classes as well as the harmonic mean of both base and unseen classes to provide a more comprehensive perspective.

\begin{figure*}
    \centering
    \includegraphics[width=1\textwidth]{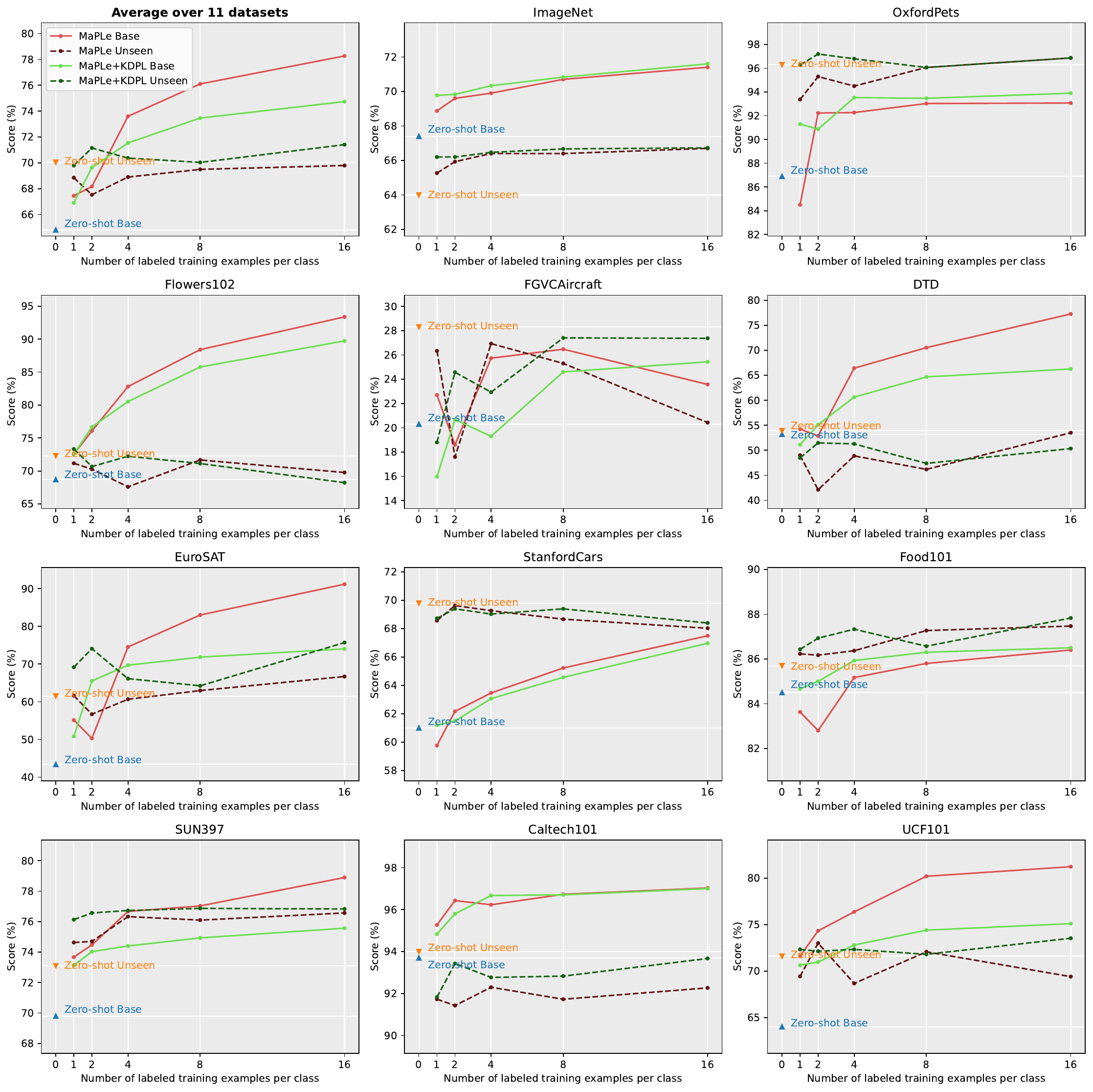}
    \caption{\textbf{Generalization to Unseen Classes}. The base-to-unseen generalization results of MaPLe+\method versus MaPLe are reported. Prompts are learned from the base classes and evaluated on the same base classes and on the unseen classes of the test set. Instead of reporting only the 16-shot performance, we include results for \mbox{1-, 2-, 4-, 8-,} and 16-shots.  
    Here we see that increasing the number of shots increases the performance of the base class but harms the performance on unseen classes. However, with our approach we can alleviate this problem, reaching an average 16-shots performance on unseen classes greater than the zero-shot reference, while the average MaPLe performance remains below the average zero-shot reference for all shots.}
    \label{fig:maple-base-16}
\end{figure*}

In Table~\ref{tab:base-to-new}, we observe that the baseline methods without \method outperform our approach in few-shot transfer learning on base classes. Thanks to the supervision label signal, the baselines can exploit the ground-truth source training classes. Without using \method, however, all the baseline methods suffer from forgetting the unseen classes. In particular, we see that training on base-shots causes a decrease in performance on the unseen classes. Certain methods in particular suffer from this phenomenon, typical of an Incremental Learning setting (e.g. in Figure~\ref{fig:maple-base-16} we see that the average performance of MaPLe on unseen classes is consistently \textit{below} the average zero-shot unseen performance). Applying our method instead exhibits a more favorable trend in zero-shot transfer learning on unseen classes, alleviating the forgetting of unseen classes. Our method is preferable in scenarios where labeled data for base classes are \textit{not} available, and improvement in base class performance is desired without compromising zero-shot transfer performance on unseen classes.

\asubsection{Class Agnostic Scenario}
\label{supp:class-agnostic}

\begin{table*}[t]
    \caption{\textbf{Class Agnostic adaptation}. Comparison between baselines and the proposed unsupervised class agnostic \CAmethod variants (highlighted in cyan). The prompt is learned in an unsupervised and class-agnostic setting on ImageNet and evaluated on the generalization and on the cross-dataset benchmark datasets. Average performance improvements are indicated in \hgreen{green}, and deterioration in \hred{red}. Notably, our proposed unsupervised and class-agnostic approach is competitive against the reference supervised and class-aware baselines. Overall, on the cross-dataset benchmark we achieve better generalization, demonstrating that using \CAmethod is a valid option in such settings.}
    \centering
    \setlength{\tabcolsep}{1pt}
    \resizebox{1\textwidth}{!}{
       \label{tab:class-agno}

        \begin{tabular}{c l cccccc ccccccccccc}
        \toprule
        & & Source & \multicolumn{5}{c}{Cross-Domain Target} & \multicolumn{5}{c}{Cross-Dataset Target}  
        \\ \cmidrule(lr){3-3} \cmidrule(lr){4-8} \cmidrule(lr){9-19}
        
        Backbone & Method & \rotatebox{80}{ImageNet} & \rotatebox{80}{ImageNet-V2} & \rotatebox{80}{ImageNet-S} & \rotatebox{80}{ImageNet-A} & \rotatebox{80}{ImageNet-R} & \rotatebox{80}{\emph{Average}} & \rotatebox{80}{OxfordPets} & \rotatebox{80}{Flowers102} & \rotatebox{80}{FGVCAircraft} & \rotatebox{80}{DTD} & \rotatebox{80}{EuroSAT} & \rotatebox{80}{StanfordCars} & \rotatebox{80}{Food101} & \rotatebox{80}{SUN397} & \rotatebox{80}{Caltech101} & \rotatebox{80}{UCF101} & \rotatebox{80}{\emph{Average}} \\ 
        \midrule
         
         & \demph{CLIP (student)} & \demph{58.20} & \demph{51.50} & \demph{33.30} & \demph{21.70} & \demph{56.00} & \demph{40.63} & \demph{83.70} & \demph{61.00} & \demph{15.60} & \demph{40.00} & \demph{24.20} & \demph{55.60} & \demph{75.20} & \demph{58.50} & \demph{86.00} & \demph{58.30} & \demph{55.81}\\
        \cmidrule(lr){2-19}
        
         & CoOp & \textbf{62.40} & \textbf{55.17} & 33.70 & 23.13 & 56.20 & 42.05 & 84.60 & 61.63 & 13.77 & 36.83 & 22.33 & 54.20 & 75.03 & 59.00 & 87.67 & 57.73 & 55.28\\

         \rowcolor{tabhighlight} \cellcolor{white} & CoOp + \CAmethod &  61.83 & 54.27 & \textbf{35.20} & \textbf{23.50} & \textbf{59.10} & \hgreen{43.02} & \textbf{86.37} & \textbf{63.47} & \textbf{14.80} & \textbf{40.23} & \textbf{30.33} & \textbf{55.40} & \textbf{77.17} & \textbf{61.60} & \textbf{88.50} & \textbf{59.03} & \hgreen{57.69}\\
        
        \cmidrule(lr){2-19}
    
         \multirow{-4}{*}{RN50} & CoCoOp & \textbf{63.07} & \textbf{55.53} & 34.77 & 23.73 & \textbf{59.47} & 43.38 & 86.93 & 63.47 & \textbf{16.27} & \textbf{40.53} & 27.20 & \textbf{55.47} & \textbf{77.73} & 61.40 & \textbf{88.07} & \textbf{60.43} & 57.75\\

         \rowcolor{tabhighlight} \cellcolor{white} & CoCoOp + \CAmethod & 61.43 & 54.53 & \textbf{35.33} & \textbf{23.77} & 59.10 & \hred{43.18} & \textbf{87.47} & \textbf{67.30} & 15.30 & 39.50 & \textbf{28.57} & 54.80 & 77.53 & \textbf{61.63} & 88.03 & 60.13 & \hgreen{58.03}\\ 
    
        \midrule
        
          & \demph{CLIP (student)} & \demph{62.00} & \demph{54.70} & \demph{40.80} & \demph{29.60} & \demph{66.00} & \demph{47.78} & \demph{85.00} & \demph{64.30} & \demph{18.20} & \demph{42.80} & \demph{38.10} & \demph{60.40} & \demph{79.10} & \demph{62.00} & \demph{91.10} & \demph{60.70} & \demph{60.17}\\
        \cmidrule(lr){2-19}
        
         & CoOp & \textbf{66.33} & \textbf{58.30} & 41.40 & \textbf{31.47} & 65.87 & 49.26 & \textbf{87.13} & 62.20 & 12.13 & 40.57 & 36.97 & 57.83 & \textbf{80.13} & 62.83 & 91.13 & \textbf{61.80} & 59.27\\
        
         \rowcolor{tabhighlight} \cellcolor{white} & CoOp + \CAmethod & 64.73 & 56.93 & \textbf{42.27} & \textbf{31.47} & \textbf{67.83} & \hgreen{49.63} & 86.57 & \textbf{62.73} & \textbf{15.00} & \textbf{42.23} & \textbf{43.93} & \textbf{58.57} & 79.47 & \textbf{65.03} & \textbf{92.30} & 61.53 & \hgreen{60.74}\\
     
        \cmidrule(lr){2-19}
    
        & VPT & \textbf{64.97} & \textbf{56.73} & \textbf{41.27} & 27.00 & 66.50 & 47.88 & \textbf{87.43} & \textbf{64.77} & \textbf{19.53} & \textbf{44.20} & 28.47 & \textbf{57.60} & \textbf{78.27} & \textbf{63.73} & 91.40 & 62.50 & 59.79\\
        
         \rowcolor{tabhighlight} \cellcolor{white} & VPT + \CAmethod & 63.73 & 55.83 & 41.10 & \textbf{27.33} & \textbf{67.30} & \hgreen{47.89} & 87.20 & 63.17 & 18.43 & 44.00 & \textbf{33.53} & 55.47 & 77.70 & 63.63 & \textbf{91.93} & \textbf{62.93} & \hgreen{59.80}\\
            
        \cmidrule(lr){2-19}
        
         \multirow{-4}{*}{ViT-B/32 } & MaPLe & \textbf{66.80} & \textbf{58.53} & \textbf{42.23} & \textbf{30.13} & 66.40 & 49.32 & \textbf{88.37} & \textbf{66.43} & \textbf{18.47} & 42.03 & 37.77 & \textbf{60.00} & \textbf{80.10} & 64.43 & 91.33 & \textbf{62.50} & 61.14\\

        \rowcolor{tabhighlight} \cellcolor{white} & MaPLe + \CAmethod & 65.23 & 57.43 & 42.07 & 30.00 & \textbf{67.63} & \hred{49.28} & 88.30 & 64.97 & 16.07 & \textbf{44.03} & \textbf{41.53} & 57.90 & 79.80 & \textbf{64.97} & \textbf{92.77} & 61.93 & \hgreen{61.23}\\  

         \cmidrule(lr){2-19}
    
        & PromptSRC & \textbf{66.33} & \textbf{58.70} & \textbf{42.97} & 32.07 & \textbf{68.93} & 50.67 & \textbf{88.13} & 65.33 & 17.17 & \textbf{43.90} & 40.50 & \textbf{60.17} & \textbf{81.23} & 65.40 & 92.37 & \textbf{63.63} & 61.78\\
        
         \rowcolor{tabhighlight} \cellcolor{white} & PromptSRC + \CAmethod & 65.00 & 57.50 & 42.90 & \textbf{32.30} & 68.77 & \hred{50.37} & 87.53 & \textbf{65.47} & \textbf{17.87} & 42.17 & \textbf{44.53} & 60.07 & 80.50 & \textbf{65.67} & \textbf{93.10} & 62.37 & \hgreen{61.93}\\
         
        \midrule
        
        ViT-H/14 & \demph{CLIP (teacher)} & \demph{82.80} & \demph{76.60} & \demph{71.10} & \demph{71.10} & \demph{91.30} & \demph{77.53} & \demph{94.80} & \demph{89.40} & \demph{63.20} & \demph{66.80} & \demph{63.30} & \demph{95.60} & \demph{93.60} & \demph{76.40} & \demph{97.90} & \demph{76.40} & \demph{81.74}\\
    
        \bottomrule
        \end{tabular}
}
\end{table*}

In the main paper we did not include all results for the class agnostic setting due to space limitations, but only reported the average domain generalization performance and the average cross-dataset performance. Here, in Table~\ref{tab:class-agno} we provide all class agnostic results comparing our proposed method (\CAmethod) against the baselines. As mentioned in the main paper, even without knowing either the labels or the training class names, our proposed unsupervised method is competitive, and sometimes even outperforms the baselines.

\end{document}